\documentclass[a4paper,fleqn]{cas-dc}

\usepackage[numbers, square, comma]{natbib}
\usepackage{float}
\usepackage[framemethod=TikZ]{mdframed}
\usepackage{xcolor}
\usepackage{algorithm}
\usepackage{algpseudocode}
\usepackage{diagbox}

\def\tsc#1{\csdef{#1}{\textsc{\lowercase{#1}}\xspace}}
\tsc{WGM}
\tsc{QE}
\tsc{EP}
\tsc{PMS}
\tsc{BEC}
\tsc{DE}

\begin{document}
\let\WriteBookmarks\relax
\def\floatpagepagefraction{1}
\def\textpagefraction{.001}
\let\printorcid\relax

\title [mode = title]{CoT-Driven Framework for Short Text Classification: Enhancing and Transferring Capabilities from Large to Smaller Model}

\author[1,2,3,4,5]{Hui Wu}[style=chinese]
\author[1,2,3]{Yuanben Zhang}[style=chinese]
\author[1,2,3]{Zhonghe Han}[style=chinese]
\author[1,2,3]{Yingyan Hou}[style=chinese]
\author[1,2,3,4]{Lei Wang}[style=chinese]
\cormark[1]
\author[1,2,3]{Siye Liu}[style=chinese]
\author[1,2,3]{Qihang Gong}[style=chinese]
\author[1,2,3]{Yunping Ge}[style=chinese]

\affiliation[1]{institute={Aerospace Information Research Institute, Chinese Academy of Sciences},
    city={Beijing},
    postcode={100190},
    country={China}}
\affiliation[2]{institute={Key Laboratory of Target Cognition and Application Technology(TCAT)},
    city={Beijing},
    postcode={100190},
    country={China}}
\affiliation[3]{institute={Key Laboratory of Network Information System Technology(NIST)},
    city={Beijing},
    postcode={100190},
    country={China}}
\affiliation[4]{institute={University of Chinese Academy of Sciences},
    city={Beijing},
    postcode={100190},
    country={China}}
\affiliation[5]{institute={School of Electronic, Electrical and Communication Engineering, University of Chinese Academy of Sciences},
    city={Beijing},
    postcode={100190},
    country={China}}

\cortext[cor1]{Corresponding author}

\begin{abstract}
Short Text Classification (STC) is crucial for processing and understanding the brief but substantial content prevalent on contemporary digital platforms. The STC encounters difficulties in grasping the semantic and syntactic intricacies, an issue that is apparent in traditional pre-trained language models. Although Graph Convolutional Networks enhance performance by integrating external knowledge bases, these methods are limited by the quality and extent of the knowledge applied. Recently, the emergence of Large Language Models (LLMs) and Chain-of-Thought (CoT) has significantly improved the performance of complex reasoning tasks. However, some studies have highlighted the limitations of their application in fundamental NLP tasks. Consequently, this study first employs CoT to investigate and enhance the capabilities of LLMs in STC tasks. We propose the Syntactic and Semantic Enrichment CoT (SSE-CoT) method, effectively decomposing the STC tasks into four distinct steps: (i) essential concept identification, (ii) common-sense knowledge retrieval, (iii) text rewriting, and (iv) classification. Furthermore, recognizing resource constraints in sectors like finance and healthcare, we then introduce the CoT-Driven Multi-Task Learning (CDMT) framework to extend these capabilities to smaller models. This framework begins by extracting rationales from LLMs and subsequently fine-tunes smaller models to optimize their performance. Extensive experimentation across six short-text benchmarks validated the efficacy of the proposed methods. In particular, SSE-CoT achieved state-of-the-art performance with substantial improvements on all datasets, particularly on the Ohsumed and TagMyNews datasets.
\end{abstract}

\begin{keywords}
Short Text Classification \sep Large Language Models \sep Chain-of-thought
\end{keywords}

\maketitle

\section{Introduction}

\begin{figure*}
    \centering
    \includegraphics[width=0.8\textwidth]{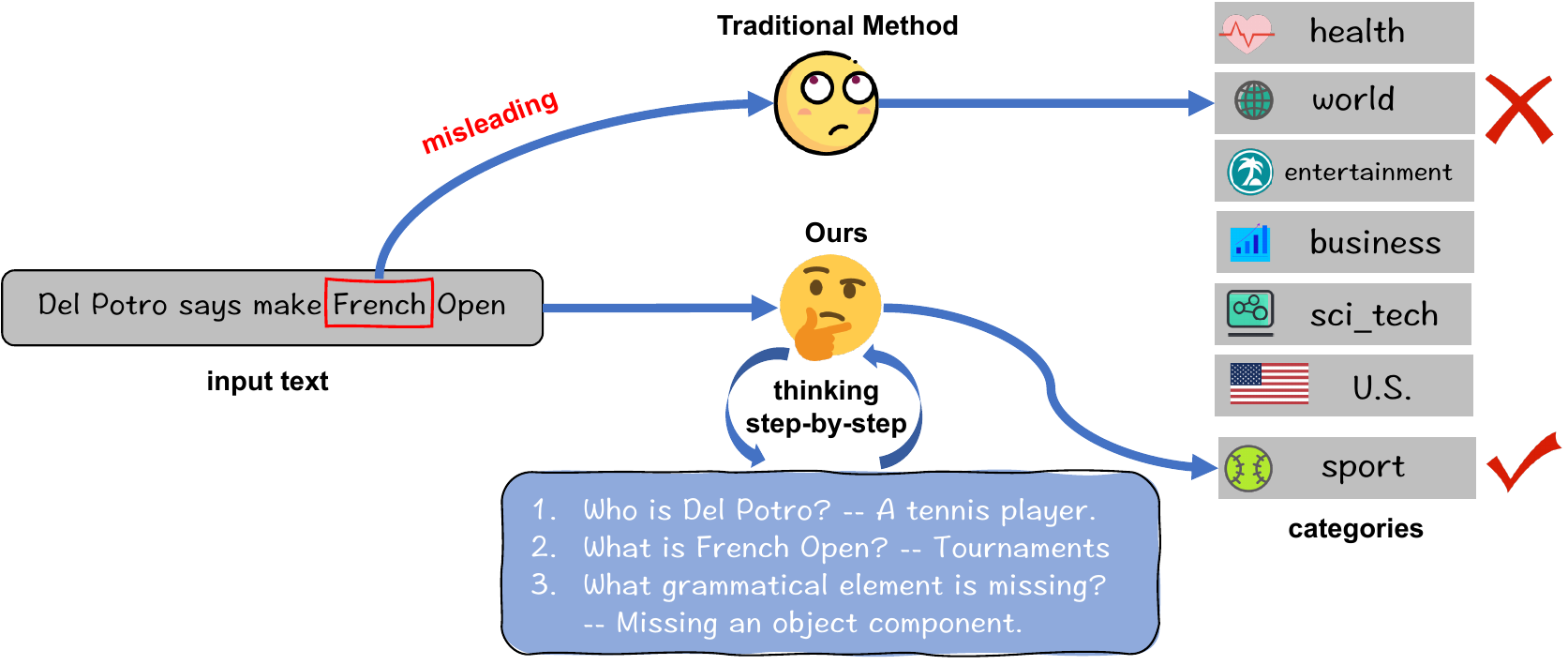}
    \caption{This diagram compares two approaches to the STC tasks. Due to misinterpretation, the traditional approach erroneously classifies the input `\textit{Del Potro says make French Open}' as `\textit{world}'. Conversely, our CoT method employs a sequential analytical process that correctly identifies `\textit{Del Potro}' as a tennis player, recognizes the `\textit{French Open}' as a tennis tournament, and detects the absence of a grammatical object in the sentence, resulting in accurate categorization under `\textit{sport}'.}
    \label{fig1}
\end{figure*}

Short texts are crucial to the contemporary flow of information, particularly with the rapid growth of the Internet\cite{phan2008learning}. They play an essential role on major social media platforms, including Twitter, TikTok, Instagram, and Weibo, where they facilitate social interaction and are integral to daily activities. As a critical task for intelligent empowerment and the application of short texts, short text classification (STC) is essential for applications such as news categorization\cite{yao2019graph}, question answering (QA)\cite{liu2019multi}, and sentiment analysis\cite{chen2019deep}. Traditional pre-trained language models (PLMs) struggle with the semantic and syntactic intricacies of STC\cite{phan2008learning, tang2015pte}. As shown in Fig \ref{fig1}, categorizing a news title such as `\textit{Del Potro says make French Open}' into the `\textit{sport}' category can be challenging, particularly without recognizing `\textit{Del Potro}' as a professional tennis player. The absence of a clear subject and predicate further hinders the model’s understanding and learning capacity. To address the challenges inherent in STC tasks, Graph Convolutional Networks (GCNs) have shown some progress by incorporating an additional knowledge base and redefining STC as a node classification issue, which mitigates the problem of limited training data. However, the effectiveness of GCNs is remains constrained by the quality and scope of the knowledge employed.
 
The NLP landscape has been revolutionized by Large Language Models (LLMs), which have achieved state-of-the-art performance in downstream tasks such as complex reasoning and QA\cite{raffel2020exploring}. Research suggests that models with larger parameters effectively function as implicit knowledge bases, offering superior integration and application of knowledge compared to traditional external knowledge bases, while demonstrating emergent abilities such as in-context learning and instruction following\cite{liu2021generated,zhou2022least}. However, some studies have highlighted limitations in applying LLMs to traditional NLP tasks\cite{wang2023gpt,wadhwa2023revisiting,gao2023exploring}. For example, in the task of named entity recognition, the performance remains significantly below that of the current best NER model\cite{wang2023gpt}. However, the application of LLMs to STC tasks remains unexplored, a gap this study aims to address. Additionally, effectively handling datasets for STC tasks, such as Ohsumed and TagMyNews, remains to be challenging. 

Our study begins by examining the challenges faced by LLMs in STC tasks. Typically, these models are pre-trained on extensive text corpora that often fail to effectively capture the semantic and syntactic nuances of short texts, thereby reducing their effectiveness. To address this issue, we utilize Chain-of-Thought (CoT) prompting\cite{wei2022chain} to improve LLMs performance in STC tasks by enabling step-by-step reasoning. Futhermore, practical challenges arises with LLMs, particularly in fileds such as finance and healthcare, where computational resources are insufficient for fine-tuning domain-specific LLMs. Smaller available models, which lack sufficient knowledge and CoT capabilities, fail to achieve comparable performance to LLMs. Accordingly, the second part of our research focuses on transferring knowledge and CoT capabilities from LLMs to smaller models, enabling them to perform effectively in resource-constrained environments.

Specifically, we propose the Semantic and Syntactic Enrichment CoT (SSE-CoT) method, designed to enhance the performance of LLMs in STC tasks. The framework divides STC tasks into four subtasks: (i) key-concept identification, which involves identifying critical words in the input text; (ii) common-sense knowledge retrieval, facilitating the acquisition of common-sense knowledge relevant to these identified keywords, thereby bridging semantic gaps in short texts; (iii) text rewriting, which reformulates the texts using this acquired knowledge to improve syntax and readability; and (iv) short-text classification, which leverages the refined texts for accurate classification.

Additionally, we introduce the CoT-Driven Multi-Task learning (CDMT) framework, aimed at enhancing smaller models\footnote{In our study, we define models with billions of parameters as large models, while those with millions of parameters are classified smaller models.} by transferring knowledge and CoT abilities from LLMs. In this framework, we first extract task-specific rationales from both SSE-CoT and Domain Augmentation CoT (DA-CoT). Subsequently, we employ multi-task learning to fine-tune the smaller models using three distinct supervision signals: the rationales from SSE-CoT and DA-CoT, as well as the ground truth. Furthermore, our Explicit Category Context Augmentation (ECCA) strategy enhances model performance by aligning predictions more closely with the ground truth.

To sum up, our contributions are as follows:
\begin{enumerate}
\itemsep=0pt
\item To the best of our knowledge, this research is the first to employ Large Language Models alongside chain-of-thought reasoning to investigate and tackle challenges in short text classification tasks.
\item We propose the Syntactic and Semantic Enrichment CoT (SSE-CoT) method to enhance the performance of LLMs for STC tasks. This approach enables LLMs to effectively decompose STC tasks and address semantic sparsity and syntactic ambiguity by breaking them into four subtasks.
\item We introduce the CoT-Driven Multi-Task learning (CDMT) framework to improve the capabilities of smaller models in STC tasks. This framework transfers knowledge and CoT abilites from LLMs to smaller models to boost their performance.
\item Comprehensive experiments were conducted using six challenging short-text benchmark datasets. The experiments confirm that our SSE-CoT method can significantly utilize LLMs for this challenging task and is superior to several novel baselines. CDMT also shows its capacity to enhance smaller models, even with limited computational resources.
\end{enumerate}

\section{Related Work}
\subsection{Text Classification}

Text classification (TC) is a fundamental task in Natural Language Processing NLP that involves assigning predefined labels to text entities\cite{li2020survey}. Historically, TC has been approached as a two-stage process: feature extraction using techniques such as term frequency-inverse document frequency\cite{blei2003latent}, followed by the application of classifiers such as support vector machines\cite{cortes1995support}. The advent of deep learning has revolutionized this approach, enabling an end-to-end methodology. Contemporary models, such as convolutional neural networks\cite{kim2014convolutional} and long short-term memory networks\cite{liu2016recurrent}, learn directly from raw data, bypass manual feature engineering, and improve classification adaptability. The introduction of BERT\cite{devlin2018bert} marked a significant milestone in deep learning, establishing itself as a prevalent choice for TC research.

Graph neural networks (GNNs) have emerged as a pivotal development in TC. Certain GNNs, such as TLGNN\cite{huang2019text}, TextING\cite{zhang2020every}, and HyperGAT\cite{ding2020more}, represent each document as a network of interconnected word nodes, effectively reframing the text classification challenge into a graph-based task. While this approach offers a novel perspective, its effectiveness diminish when dealing with limited labeled data. Conversely, models such as TextGCN\cite{yao2019graph} and TensorGCN\cite{liu2020tensor} adopt a broader perspective, framing the classification task within corpus-level graphs, where both individual words and entire texts are presented as nodes. These models use node classification techniques to identify and classify unlabeled textual elements. However, these models often struggle when handling concise textual data or datasets with limited contextual richness.

Recently, the introduction of ChatGPT has revolutionized the field. Numerous studies have explored the application of Large Language Models in TC tasks. Some studies have leveraged ChatGPT for automated genre recognition to streamline the text classification process through its zero-shot classification abilities\cite{kuzman2023chatgpt} and evaluated the capacity of ChatGPT for text classification within affective computing by employing it in tasks such as personality prediction, sentiment analysis, and suicidal ideation detection tasks\cite{amin2023will}.

\subsection{Short Text Classification}

Short Text Classification (STC) has consistently posed significant challenges in the field of NLP, presenting unique complexities that sharply contrast with those of traditional TC tasks\cite{wang2016semantic,li2020survey,wang2015semantic}. First, the brevity of short texts inherently limits its semantic and syntactic richness\cite{phan2008learning,tang2015pte}. Researchers have explored methods to enhance the expressiveness of short texts by integrating additional information\cite{xu2020incorporating,liu2022combining,chen2011short}. Common approaches include concepts from external knowledge bases such as \cite{wang2017combining} and latent topics uncovered within the corpus\cite{zhang2016improving, zeng2018topic}. Second, STC often encounters the challenge of sparsely labeled data in practical applications\cite{chen2019deep,yang2022survey}, which exacerbates increasing the task complexity. A prevalent strategy to mitigate this involves employing graph-based methods, which not only supplement additional information but also offset the paucity of label data, as evidenced by the approach adopted by \cite{linmei2019heterogeneous}, constructing a corpus-level graph that models latent topics, entities, and documents jointly, where the entities are words linked to knowledge graphs. SHINE\cite{wang2021hierarchical} introduces a hierarchically organized heterogeneous corpus-level graph, comprising word-level and document-level graphs, to fully exploit interactions between nodes of the same type and capture similarities between short texts. ST-Text-GCN\cite{cui2022self} uses a self-training method for keyword extraction, effectively leveraging limited labeled texts and a large number of unlabeled texts.  

Despite significant advancements in graph-based methodologies, these approaches exhibit clear limitations in certain practical scenarios. One of the primary drawbacks is the need to retrain the entire model to incorporate new test samples, which can be both computationally intensive and time-consuming. As a result, there has been a growing interest in inductive reasoning approaches, which eliminate the need for retraining by integrating new samples directly with existing training and unlabeled data. Innovations such as HGAT-inductive\cite{yang2021hgat}, propose a novel framework for inductively linking each new sample to an existing corpus, thereby facilitating dynamic learning. SimpleSTC\cite{zheng2022simplified} adopts a word-only approach to address the inductive STC problem accurately. Nevertheless, while these inductive methods offer increased flexibility and efficiency, they may necessitate a trade-off in terms of precision as they rely on the assumption that new samples share similar characteristics with the existing corpus.

\subsection{Chain-of-thought in LLMs}
Large Language Models (LLMs), such as \cite{ouyang2022training,touvron2023llama,du2021glm} have gained significant attention for their advancements in dialogue systems and potential across various applications\cite{liu2021generated,zhou2022least}. The Chain of Thought (CoT) strategy improves the reasoning capabilities of LLMs by employing a structured, step-by-step process suitable for complex reasoning tasks\cite{wei2022chain,wang2022self,zelikman2022star}.

Recent studies have explored the use of the CoT approach to further enhance the capabilities of LLMs.  For instance, \cite{wei2022chain} improved learning and reasoning in LLMs by manually constructing CoT prompts with specific examples to facilitate the analysis of complex problems. \cite{gao2023pal} incorporated programming languages as annotated rationales in their PAL method, converting problem-solving into executable Python programs and demonstrating CoT's utility in programmed tasks.  Additionally, \cite{fei2023reasoning} employed the CoT strategy with a three-step prompting principle, effectively inferring the latent intent of opinions to address Implicit Sentiment Analysis. \cite{wadhwa2023revisiting} applied the CoT strategy to fine-tuning GPT-3 and Flan-T5 enhancing the processing of complex semantic relationships in Relation Extraction tasks. \cite{zou2023meta} introduced GeM-CoT, a generalizable CoT mechanism designed to enhance performance and generalization across diverse mixed-task scenarios.

Various CoT modifications have been proposed to optimize reasoning processes. The Tree of Thought \cite{yao2023tree} linearizes reasoning for retrospective analysis, while the Graph of Thought \cite{besta2023graph} restructures it into a directed acyclic graph textcolor{red}{to enhance} navigation. The Array of Thoughts \cite{sel2023algorithm} maintains a dynamic context chain  to minimize repetitive querying. Although LLMs have been explored for certain NLP basic tasks\cite{wang2023gpt,wadhwa2023revisiting,gao2023exploring}, their application in STC tasks using Chain-of-Thought strategy remains unexplored until our study, motivating the focus of our study. Therefore, this paper explores the integration of CoT with LLMs for addressing STC tasks.

\subsection{Knowledge Distillation in LLMs}
In recent years, LLMs have undergone significant evolution, with their parameter sizes have expanded significantly. Research suggests that models with larger parameters usually achieve better performance\cite{ouyang2022training}. However, the increasing scale of these models presents challenges, such as higher deployment costs and decreased training efficiency. Knowledge distillation \cite{hinton2015distilling} is a method that compresses models by transferring knowledge from a large, well-trained teacher model to a smaller student model.

Traditional knowledge distillation methods using LLMs, such as GKD\cite{agarwal2023gkd} and MINILLM\cite{gu2023knowledge}, primarily focus on distilling specific outputs from the teacher model, like classification labels. These methods directly transfer direct knowledge from the teacher to the student model, simplifying complex concepts to facilitate easier learning for the smaller model. In contrast, the approach presented in this paper emphasizes distilling the teacher model's thought process in response to inputs, capturing the reasoning underpinning its decisions. Recent advancements, notably MT-COT\cite{li2022explanations}, integrate nuanced elements of the teacher's thought process into the student model’s training. This enhances the student's ability to manage complex tasks and multitask through strategically crafted prompts. Furthermore, SOCRATIC COT \cite{shridhar2022distilling} distills reasoning capabilities from the teacher model into the student model to solve complex problems.

\section{Method}

In section \ref{section3.1}, we define the STC tasks and then reiterate its primary challenges. Section \ref{section3.2} introduces SSE-CoT, a specialized CoT method designed for LLMs to better address the STC tasks. Finally, we detail the CoT-Driven Multi-Task learning method developed to enable smaller models to tackle the STC tasks effectively in section \ref{section3.3}.

\subsection{Task Description}\label{section3.1}
Given a dataset $D$ consisting of N short texts, the STC tasks aim to classify text $x_i\in D$ as a relevant label $l$ from a predefined set of labels $L$.
Semantic and syntactic challenges are inherent in STC tasks owing to their concise and informal nature. State-of-the-art methods based on GCN are constrained by their dependence on an external knowledge base. To address these limitations, we propose distinct methods for LLMs and smaller models that target the issues of limited context and dependency on external knowledge.

\subsection{Semantic  and Syntactic Enrichment CoT in LLMs}\label{section3.2}
To enhance the performance of LLMs in handling STC tasks, this study introduces the Semantic and Syntactic Enrichment CoT (SSE-CoT). Unlike traditional methods that employ single-step prompts, SSE-CoT employs a multi-step reasoning process specifically designed to enhance both the semantic and syntactic understanding of short texts within LLMs.

\begin{itemize}
\item \textbf{Semantic elements} are concerned with the meanings that words, phrases, and sentences convey through the short texts. They involve understanding the implications, nuances, and contextual uses of language within a narrative.
\item \textbf{Syntactic elements} refer to the arrangement of words and phrases to create well-formed sentences according to the rules of grammar. This includes the structure of sentences, the correct use of grammatical rules, and the logical relationships between different parts of the text.
\end{itemize}

\begin{figure*}
    \centering
    \includegraphics[width=1.0\textwidth]{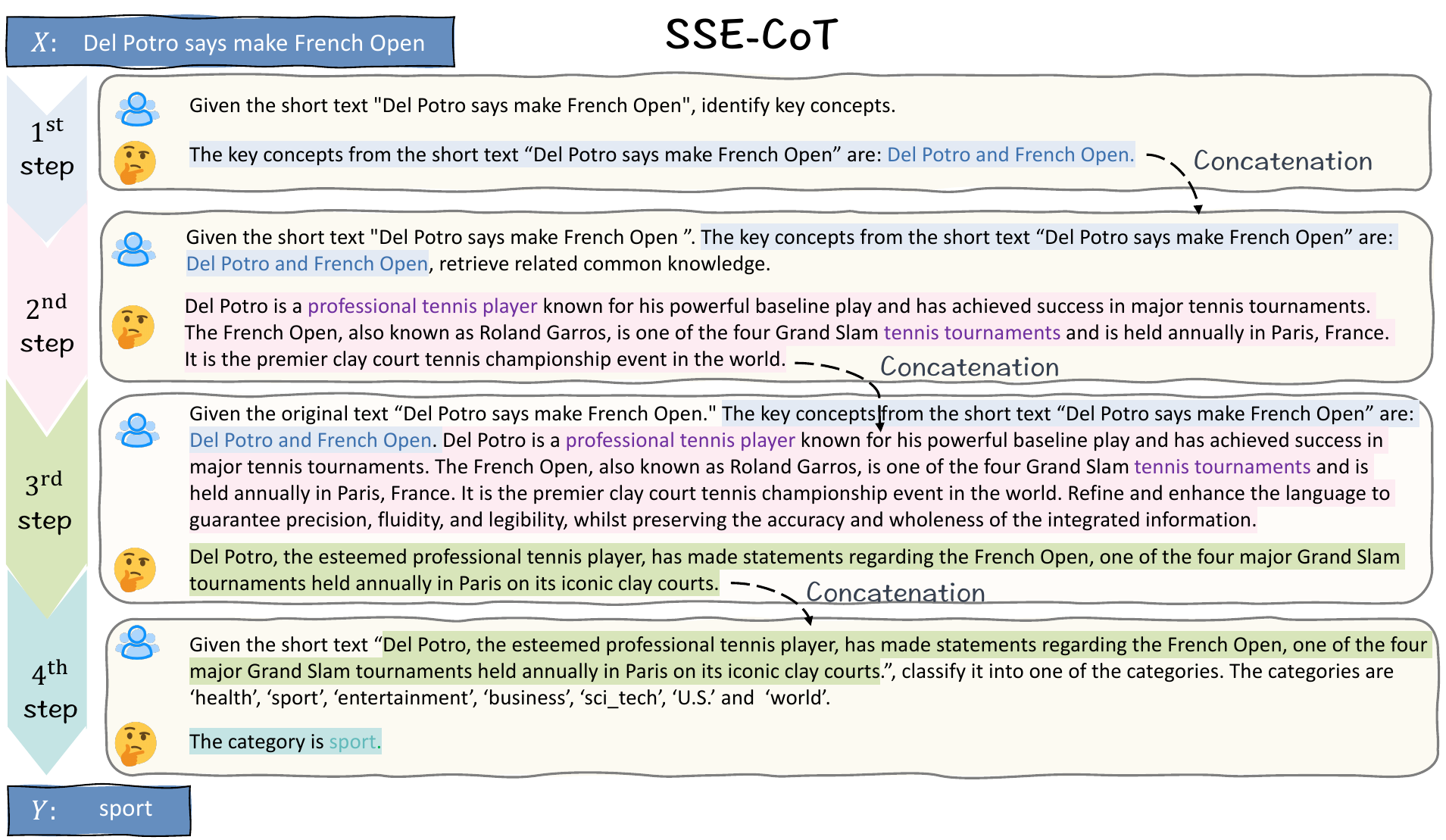}
    \caption{This diagram presents the \textbf{Semantic  and Syntactic Enrichment CoT (SSE-CoT)}, as applied to the short text `\textit{Del Potro says make French Open}'. It begins by identifying key concepts, `\textit{Del Potro}' and the `\textit{French Open}', then combines them to contextualize `\textit{Del Potro}' as a tennis player and the `\textit{French Open}' as a major tournament. The third step refines this information for accuracy and integration. Finally, the process classifies the outcome under `\textit{sport}'. The framework offers a novel solution that effectively addresses STC tasks challenges.}
    \label{fig2}
\end{figure*}

Our SSE-CoT method comprises four distinct steps, as shown in Fig \ref{fig2}. First, it identifies key concepts and terms to establish a foundational understanding. The second step, semantic enrichment, deepens the model's comprehension of these concepts and their relationships. This is followed by syntactic enrichment, which refines the grammar and structure of the text. The final step integrates these enhancements, enabling the model to make predictions. Specifically, we conduct our four-step prompts as follows.
\vspace{0.5em} 

\noindent
\textbf{Step 1. Key Concept Identification}
\vspace{0.5em} 

We first ask LLM to identify relevant concepts using a specified template:
\begin{mdframed}[
    outerlinewidth=0.5pt,
    roundcorner=5pt,
    innertopmargin=5pt,
    innerbottommargin=5pt,
    innerrightmargin=5pt,
    innerleftmargin=5pt,
    backgroundcolor=gray!7,
    linecolor=black,
    align=center,
    userdefinedwidth=0.5\textwidth
]
\centering
\textsf{$C^1_1$[Given the short text $x_i$], identify key concepts.}
\end{mdframed}

Here, $C^1_1$ is the context of the first step, and the following content is the instruction $I^1_1$. This stage was designed to focus the model on essential content in preparation for the next steps. The process can be formally expressed as:
\begin{align}
    K^1 = f_{\mathrm{identify}}(C^1_1,I^1_1)
\end{align}
where $f_{\mathrm{identify}}$ denotes a function that captures the capability of the models to extract concepts.
\vspace{0.5em} 

\noindent
\textbf{Step 2. Common-sense Knowledge Retrieval} 
\vspace{0.5em} 

With the fundamental concepts $K^1$ identified in Step 1, this step involves retrieving the associated common-sense knowledge from the inherent knowledge base of the LLM using the following template:
\begin{mdframed}[
    outerlinewidth=0.5pt,
    roundcorner=5pt,
    innertopmargin=5pt,
    innerbottommargin=5pt,
    innerrightmargin=5pt,
    innerleftmargin=5pt,
    backgroundcolor=gray!7,
    linecolor=black,
    align=center,
    userdefinedwidth=0.5\textwidth
]
\centering
\textsf{$C^1_2$[$C^1_1$,$K^1$], retrieve related common knowledge.}
\end{mdframed}

In this phase, concatenate $C^1_1$ and $K^1$ to form the context, and use the following content $I^1_2$ as the retrieval directive:
\begin{align}
    S = f_{\mathrm{retrieve}}(C^1_2,I^1_2)
\end{align}
here, $f_{\mathrm{retrieve}}$ is a function enabling the model to recall pertinent information from an internal knowledge repository. Knowledge retrieval mitigates the semantic gap in STC tasks because of its brevity, and facilitates the provision of contextually rich and comprehensive responses.
\vspace{0.5em} 

\noindent
\textbf{Step 3. Text Rewriting} 
\vspace{0.5em} 

Following the retrieval of pertinent common-sense knowledge, this step entails assimilating $S$ into a cohesive and polished short text. The context is formed by concatenating $C^1_2$ and $S$, where $I^1_3$ serves as the modification directive.
\begin{mdframed}[
    outerlinewidth=0.5pt,
    roundcorner=5pt,
    innertopmargin=5pt,
    innerbottommargin=5pt,
    innerrightmargin=5pt,
    innerleftmargin=5pt,
    backgroundcolor=gray!7,
    linecolor=black,
    align=center,
    userdefinedwidth=0.5\textwidth
]
\textsf{$C^1_3$[$C^1_2$,$S$]. Refine and enhance the language to guarantee precision, fluidity, and legibility, whilst preserving the accuracy and wholeness of the integrated information.}
\end{mdframed}

Integration, represented by function $g$, is essential for converting raw data into an easily understandable format. It overcomes the syntactic constraints of short texts, producing a structured output $R$ that simplifies comprehension and subsequent categorization by LLM. The process can be formally expressed as:
\begin{align}
    R = g(C^1_3,I^1_3)
\end{align}

\vspace{0.5em} 
\noindent
\textbf{Step 4. Short Text Classification}
\vspace{0.5em} 

After integrating common-sense knowledge and refining the short text $x_i$ as $R$, we prompted the LLM with instruction $I^1_4$ to generate the final predicted label.
\begin{mdframed}[
    outerlinewidth=0.5pt,
    roundcorner=5pt,
    innertopmargin=5pt,
    innerbottommargin=5pt,
    innerrightmargin=5pt,
    innerleftmargin=5pt,
    backgroundcolor=gray!7,
    linecolor=black,
    align=center,
    userdefinedwidth=0.5\textwidth
]
\textsf{Given the short text $R$. classify it into one of the categories. The categories are `health’, `sport’, `entertainment’, `business’, `sci\_tech’, `U.S.’ and `world’.}
\end{mdframed}

The process can be formally expressed as:
\begin{align}
    \hat{y_i} = \mathrm{argmax} p(y|R,I_4)
\end{align}

The label with the highest output probability is designated as the predicted label $\hat{y_i}$.

\subsection{CoT-Driven Multi-Task learning for Smaller Models}\label{section3.3}

\begin{figure*}
    \centering
    \includegraphics[width=1.0\textwidth]{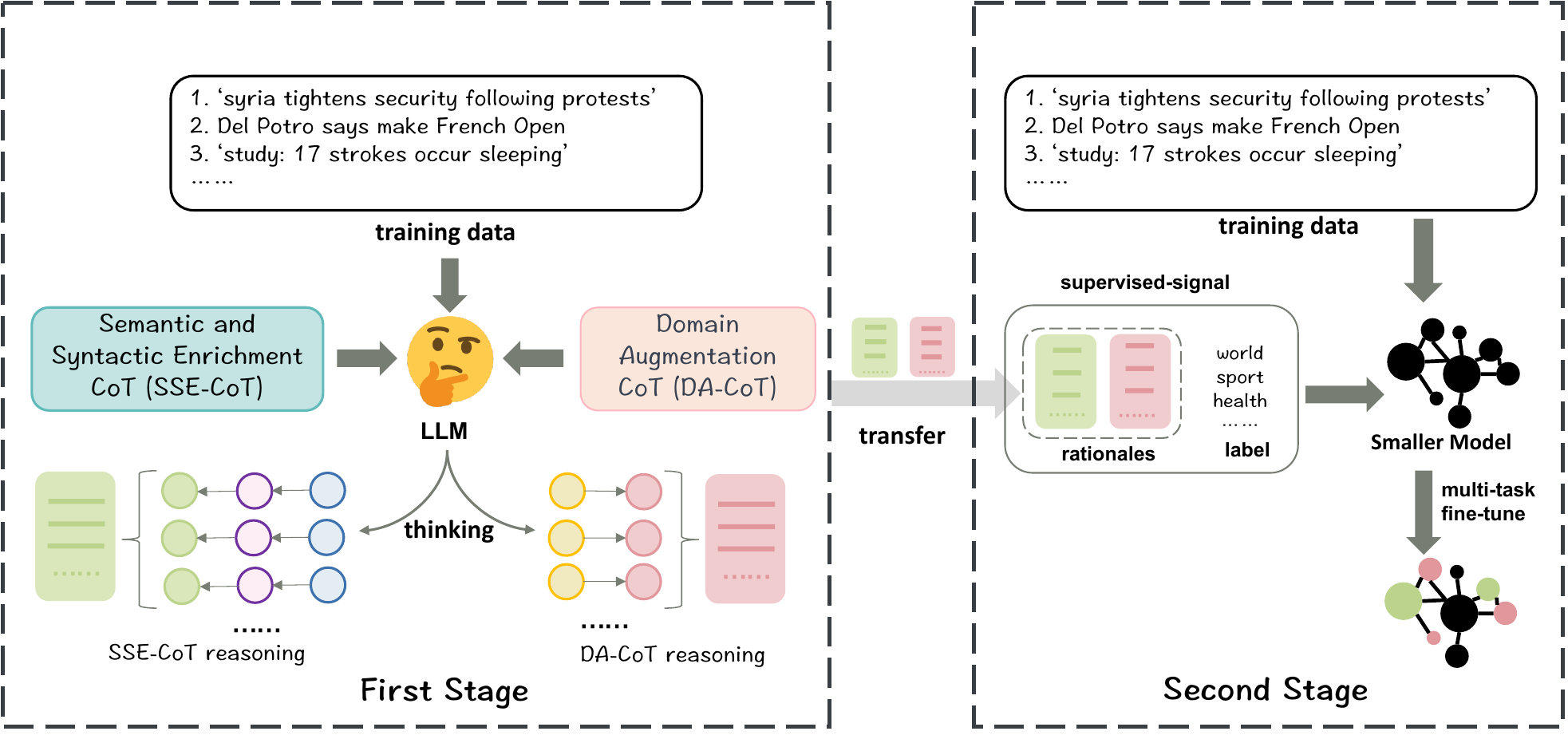}
    \caption{Overview of the CDMT method. In the first stage, the framework employs SSE-CoT and DA-CoT to prompt LLM with training data for rationale generation. In the second stage, the generated rationales guide the training of a smaller, specialized model. This stage involves multi-task fine-tuning that incorporates a supervised signal, which includes a label and two distinct rationales derived from SSE-CoT and DA-CoT reasoning.}
    \label{fig3}
\end{figure*}

We propose the CoT-Driven Multi-Task learning (CDMT)  method for STC tasks using smaller models. The architecture of this framework is depicted in Fig \ref{fig3}. Our framework comprises two stages, described below.

In the initial stage, we employed two specialized CoT prompt processes to generate rationales from LLM. The SSE-CoT enhances the clarity and coherence of short texts by addressing their inherent limitations.  We define a rationale as a chain of reasoning processes generated by LLMs in this paper. Concurrently, Domain Augmentation CoT (DA-CoT) enriches the textual context by incorporating domain-specific knowledge. In the second stage, knowledge transfer occurs from the LLM to a smaller model through a multi-task learning strategy. The smaller model was trained to predict labels and generate common-sense and domain-specific rationales. This integrated training approach improves the reasoning capabilities of the model and strengthens its ability to classify short texts with precision and depth.

We introduce rationale generation in section \ref{s.3.3.1}, followed by an overview of Explicit Category Context Augmentation in section \ref{s.3.3.2}, providing a direct and efficient prompt. Finally, we present our multi-task learning strategy in section \ref{s.3.3.3}.

\subsubsection{Rationale generation}\label{s.3.3.1}

\begin{figure*}
    \centering
    \includegraphics[width=1\textwidth]{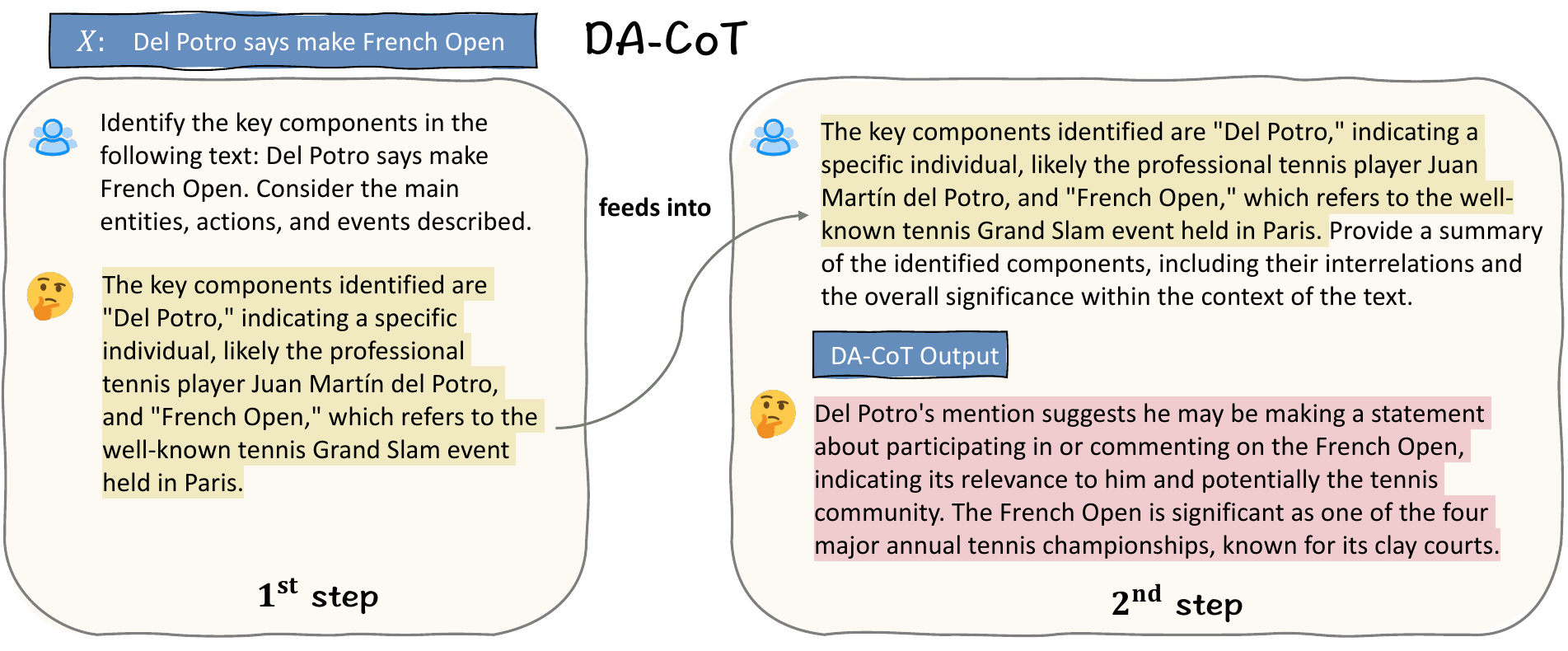}
    \caption{The figure depicts the two-phase procedure of the DA-CoT method employed in the snippets domain. Initially, the method discerns essential text elements, including primary entities, actions, and events. Subsequently, it synthesizes the interconnections and collective importance of these elements, enhancing comprehension of their pertinence and consequences in the context of the text.}
    \label{fig4}
\end{figure*}

In our study, we introduce two distinct CoTs to generate rationales. First, SSE-CoT introduced in Section \ref{section3.2},SSE-CoT is specifically designed for STC tasks to address the unique characteristics of short texts directly. Second, DA-CoT aims to enhance the performance of smaller models by transferring more domain knowledge.  This approach follows a two-step reasoning process, as shown in Figure \ref{fig4}.  Specifically, we conduct DA-CoT prompts as follows.

\vspace{0.5em} 
\noindent
\textbf{Step 1. Key Concept Identification} 
\vspace{0.5em} 

In the first step, we used a designated template to query the LLM, capitalizing on its capacity to pinpoint pertinent concepts. Unlike the first step of the SSE-CoT, the DA-CoT incorporates domain-specific cue words. For example, in the news domain illustrated in Fig \ref{fig4}, it is essential to include critical entities, actions, and events. Details of the other domains are provided in Appendix \ref{appendixA}. $C^2_1$ comprises the context of the first step, and the following content constitutes the reasoning instruction $I^2_1$ for the first step.

\begin{mdframed}[
    outerlinewidth=0.5pt,
    roundcorner=5pt,
    innertopmargin=5pt,
    innerbottommargin=5pt,
    innerrightmargin=5pt,
    innerleftmargin=5pt,
    backgroundcolor=gray!7,
    linecolor=black,
    align=center,
    userdefinedwidth=0.5\textwidth
]
\textsf{$C^2_1$[Given the short text $x_i$], identify the key components, consider the main entities, actions, and events described.}
\end{mdframed}

\begin{align}
    K^2 = f_{\mathrm{identify}}(C^2_1, I^2_1)
\end{align}
Here, $K^2$are the key snippet concepts extracted from $x_i$ using $f_{\mathrm{identify}}$ representing the model's capability to identify and understand key snippet concepts and terminologies from the text.

\vspace{0.5em} 
\noindent
\textbf{Step 2. Domain Knowledge Retrieval} 
\vspace{0.5em} 

After establishing foundational concepts in the initial phase, this step prompts the LLM to apply domain-specific terminology and deeper analytical perspectives to its outputs.
\begin{mdframed}[
    outerlinewidth=0.5pt,
    roundcorner=5pt,
    innertopmargin=5pt,
    innerbottommargin=5pt,
    innerrightmargin=5pt,
    innerleftmargin=5pt,
    backgroundcolor=gray!7,
    linecolor=black,
    align=center,
    userdefinedwidth=0.5\textwidth
]  
\textsf{$C^2_2$[$C^2_1$,$K^2$]. Provide a summary of the identified components, including their interrelations and the overall significance within the context of the text.}
\end{mdframed}
\begin{align}
    O = f_{\mathrm{enrich}}(C^2_2,I^2_2)
\end{align}
where $O$ represents the enriched knowledge retrieved or generated by the function $f_{\mathrm{enrich}}$, integrating the identified concepts with in-depth, domain-specific information.

The rationales generated by SSE-CoT directly influence the classification outcomes, whereas those from DA-CoT primarily enhance performance implicitly. Thus, it is necessary to filter and verify the rationales produced by SSE-CoT. Following the CROP method \cite{li2022explanations}, We input $x_i$ with SSE-CoT template into the LLM to obtain the intermediate explanation $R$ and the predicted label $\hat{y_i}$. We accept $r_i = R$ only if $y_i = \hat{y_i}$. If they do not match, we concatenate $x_i$ with its true label $y_i$ to generate $R$ without applying any filter.

\subsubsection{Explicit Category Context Augmentation}\label{s.3.3.2}
In contrast to encoder-only models such as BERT, the smaller model utilized in our study adopts an encoder-decoder architecture. This generative approach requires the prompt enhancement of the comprehension and response of the model to the task. To circumvent the unpredictability inherent in manually crafted prompts, we introduce the Explicit Category Context Augmentation (ECCA) method, which eliminates the requirement for manually crafted task-specific prompts and enriches text representation, leading to a more accurate model classification.

In the ECCA method, the original input text $x_i$ is augmented with category labels $L$ using an injection function to form an enhanced input $x_i'$, which the model subsequently uses for classification. This process was designed to infuse label-specific semantic cues into a model classification task.
\begin{equation}
x_i' = \text{inject}(x_i, L)
\end{equation}
The injection can be executed as a simple concatenation, where \# represents the concatenation operation and $l_j,j\in (1,2,\cdots,m) $ denotes each label. 
\begin{equation}
x_i' = l_{1} \oplus l_{2} \oplus \ldots \oplus l_{m} \oplus x_i
\end{equation}

\subsubsection{Multi-task learning}\label{s.3.3.3}

Several methods exist for integrating rationale into the training processes of downstream models. The direct method uses rationale as an additional input. However, this approach requires that an LLM generate a rationale before the smaller model makes a prediction. Therefore, we adopt a multi-task learning framework to enhance the link between the input $x_i$ and desired output $y_i$.

The primary task is to predict the correct category label $\hat{y_i}$ from augmented input $x_i'$. The secondary task involves processing the original text $x_i$ to output a rationale $\hat{r_i}$, with $r_i$ defined in Section \ref{s.3.3.1} serving as ground truth. Similarly, the tertiary task processes the original text $x_i$ to generate the domain-specific rationale $\hat{o_i}$, where $o_i\in O$ is the ground truth. This multi-task setup aims to predict the category label directly and generate rationales that provide interpretability and context to the model's decisions, leveraging the rationales from the teacher model as supplementary guidance. Thus, the loss function encompasses the following terms for each task:
\begin{equation}
L = L_{\text{label}} + \lambda_1 L_{\text{SSE}} + \lambda_2 L_{\text{DA}}
\end{equation}
The weights \( \lambda_1 \) and \( \lambda_2 \) balance the influences of the secondary and tertiary tasks, ensuring that any single task does not dominate the model's training. The calculation method for each loss function is as follows:
\begin{equation}
L_{\text{label}} = \frac{1}{N} \sum_{i=1}^{N} \ell(f_s(x_i'), \hat{y_i})
\end{equation}
\begin{equation}
L_{\text{SSE}} = \frac{1}{N} \sum_{i=1}^{N} \ell(f_s(x_i),r_i)
\end{equation}
\begin{equation}
L_{\text{DA}} = \frac{1}{N} \sum_{i=1}^{N} \ell(f_s(x_i), o_i)
\end{equation}
Here, $f_s$ represents the smaller model, and $\ell$ denotes the cross-entropy loss between the predicted and target tokens. The estimated rationales are unnecessary during testing, thereby obviating the need for LLM at that stage.

\begin{itemize}
    \item The primary loss function, $L_{\text{label}}$, ensures that the model correctly classifies the short texts into their respective categories.
    \item The second, $L_{\text{SSE}}$, relates to the SSE-CoT rationale generation, ensuring that the model not only performs well on classification but also aligns its reasoning with how the SSE-CoT enhances short text understanding.
    \item The third, $L_{\text{DA}}$, associated with the DA-CoT, ensures that domain-specific knowledge is incorporated effectively.
\end{itemize}

Optimizing short text classification, text refinement, and related knowledge prediction within a multi-task framework effectively utilizes the induced knowledge of LLMs. By incorporating moderate inductive biases into the parameter space, this approach enhances the generalization performance and robustness of smaller models.

\begin{algorithm}
\caption{CDMT Framework}
\begin{algorithmic}[1]

\State \textbf{Input:} Samples $X = \{x_1, x_2, \dots, x_n\}$ and labels $Y = \{y_1, y_2, \dots, y_n\}$

\Comment{First Stage: Rationale Generation with LLMs}
\For{each $(x_i,y_i)$ in $(X, Y)$}
    \State Generate $R, \hat{y}_i$ using SSE-CoT with $x_i$
    \If{$\hat{y}_i =y_i$}
        \State Set $r_i = R$
    \Else
        \State Concatenate $x_i$ with true label $y_i$
        \State Regenerate $R, \hat{y}_i$ using SSE-CoT with $x_i,y_i$
        \State Set $r_i = R$ 
    \EndIf
    \State Generate $O$ using DA-CoT with $x_i$
    \State Set $o_i = O$
\EndFor

\Comment{Second Stage: Fine-tuning Smaller Model}
\State Initialize smaller model
\For{each $(x_i, r_i, y_i, o_i)$}
    \State Prepare $x'_i = l_1 \oplus l_2 \oplus \ldots \oplus l_m \oplus x_i$
    \State \textbf{Train smaller model using multi-task learning}:
        \State \quad - Compute loss $L_{\text{label}}$ based on $\ell(f_s(x'_i), y_i)$
        \State \quad - Compute loss $L_{\text{SSE}}$ based on $\ell(f_s(x_i), r_i)$
        \State \quad - Compute loss $L_{\text{DA}}$ based on $\ell(f_s(x_i), O_i)$
        \State \quad - Update model by minimizing $L = L_{\text{label}} + \lambda_1 L_{\text{SSE}} + \lambda_2 L_{\text{DA}}$
\EndFor

\State \textbf{Output:} Fine-tuned smaller model
\end{algorithmic}
\end{algorithm}

\section{Experiment}
\subsection{Datasets}
To ensure a thorough and unbiased evaluation, we conducted extensive experiments on six widely recognized benchmark short-text datasets: MR, Snippets, Ohsumed, StackOverflow, TagMyNews, and AGNews. Following \cite{wang2021hierarchical}, we randomly sampled 40 labeled short texts from each class, where half formed the training set, and the other half formed the validation set. Table \ref{table1} provides detailed information regarding the datasets. We describe the datasets in more detail further below.

\begin{table}
    \centering
    \resizebox{0.45\textwidth}{!}{
    \begin{tabular}{ccccc}
    \toprule
         &\raisebox{0.5ex}{\tiny\#}texts & avg.length & \raisebox{0.5ex}{\tiny\#}classes & \raisebox{0.5ex}{\tiny\#}train(radio)\\
    \midrule
        MR
        & 10662 & 12.1 & 2 & 40(0.38\%)\\
        Snippets & 12340 & 17.4 & 8 & 160(1.30\%)\\
        Ohsumed & 7400 & 8.65 & 23 & 460(6.22\%)\\
        StackOverflow & 20000 & 5.7 & 20 & 400(2.00\%) \\
        TagMyNews & 32605 & 6.1 & 7 & 140(0.43\%)\\
        AGNews & 20000 & 27.8 & 4 & 80(0.4\%)\\
    \bottomrule
    \end{tabular}}
    \caption{Summary of short text datasets used.}
    \label{table1}
\end{table}

\begin{enumerate}
\itemsep=0pt
\item \textbf{MR}\footnote{\url{https://www.cs.cornell.edu/people/pabo/movie-review-data/}}: introduced by \cite{pangb2005exploitingclassrelationshipsforsentimentcate}, consists of one-sentence movie reviews annotated as either positive or negative for sentiment analysis.
\item \textbf{Snippets}\footnote{\url{https://github.com/jacoxu/STC2/tree/master}}: introduced by \cite{phan2008learning}, are snippets derived from Google search engine results.
\item \textbf{Ohsumed}\footnote{\url{https://github.com/yao8839836/text_gcn}}: introduced by \cite{hersh1994ohsumed}, this dataset focused on classifying cardiovascular diseases. In this study, we utilized the subset defined by \cite{linmei2019heterogeneous}, which concentrates on the classification of short texts using only the titles of single-label documents.
\item \textbf{StackOverflow}\footnote{\url{ https://github.com/jacoxu/StackOverflow/blob/master/}}: introduced by \cite{yao2019graph}, contains 20 distinct types of question titles, derived from the StackOverflow data within the IT Q\&A community.
\item \textbf{TagMyNews}\footnote{\url{https://github.com/cskarthik93/TagMyNews-Classification}}: introduced by \cite{linmei2019heterogeneous}, comprises English news headlines sourced from the Really Simple Syndication (RSS) feeds.
\item \textbf{AGNews}\footnote{\url{https://github.com/mhjabreel/CharCNN}}: introduced by \cite{zhang2015character}, was the source of our dataset. In this study, we randomly selected 20,000 articles from the collection.
\end{enumerate} 

\subsection{Experimental Setting}

We selected LLaMA2-13B\cite{touvron2023llama} as the foundational model for our SSE-CoT method and chose the LLaMA2-13B and Flan-T5-Large\cite{chung2022scaling} models to represent the LLM and smaller model in the CDMT method, respectively. In the SSE-CoT method, the LLaMA2-13B employs efficient parameter fine-tuning with Low-Rank Adaptation (LoRA)\cite{hu2021lora}, which is conducted with a batch size of 10 across five epochs. LoRA maintains the weights of pretrained LMs while introducing trainable rank decomposition matrices into each transformer layer, making it feasible to fine-tune larger LMs with fewer computational resources\footnote{In our experiment, trainable parameters only account for 0.24\% of the entire LLaMA2-13B parameters}. Conversely, in the CDMT, Flan-T5-Large has fully fine-tuned parameters with batch sizes of 5 and 10 epochs. All experiments were conducted using three A100s and five V100s.

\subsection{Evaluation}
To assess the efficacy of the proposed approach, we selected two principal evaluation metrics: accuracy (denoted as \textbf{ACC}) and macro-averaged F1 score (denoted as \textbf{F1}). Accuracy provides a straightforward measure of the overall accuracy of the model predictions. Conversely, the macro-averaged F1 score is crucial for imbalanced datasets because it uniformly weighs precision and recalls across all classes, ensuring fair evaluation even with limited samples in some categories.

\subsection{Compared Methods}
The baselines can be divided into three primary categories: \textbf{ group (A)}, pre-trained language models; \textbf{ group (B)}, GCN-based Models; and \textbf{ group (C)}, Large Language Models. Each group is discussed below.

\noindent
\textbf{Group (A). Pre-trained Language Models}

Pre-trained Language Models (PLMs) have garnered considerable attention in the field of NLP, frequently serving as tools for text classification, among other basic tasks. In this study, we chose three established methods for comparative analysis.
\begin{enumerate}
\itemsep=0pt
\item BERT\footnote{\url{https://tfhub.dev/tensorflow/bert_en_uncased_L-12_H-768_A-12/4}}\cite{devlin2018bert}, pre-trained on extensive corpora, is further fine-tuned using a linear classifier for short-text classification. Each document can be represented by either the average of its word embeddings (denoted by \textbf{-avg}) or the embedding of a CLS token (denoted by \textbf{-CLS}).
\item RoBERTa\footnote{\url{https://github.com/facebookresearch/fairseq}}\cite{liu2019roberta} is a modified and optimized version of BERT that is pre-trained on large amounts of text using adjusted training approaches.
\end{enumerate} 

\noindent
\textbf{Group (B). GCN-based models}

Recently, methods based on Graph Convolutional Neural Networks (GCNs) have achieved outstanding results in STC tasks. Five classical models are used in this study.
\begin{enumerate}
\itemsep=0pt
\item HGAT-inductive\footnote{\url{https://github.com/BUPT-GAMMA/HGAT}}\cite{yang2021hgat} advocates connecting each new sample to the existing training and unlabeled data in the corpus.
\item SimpleSTC\footnote{\url{https://github.com/tata1661/SimpleSTC-EMNLP22}}\cite{zheng2022simplified} constructs a basic word graph of common words using an additional corpus to address STC tasks via inductive learning.
\item ST-Text-GCN\footnote{\url{https://github.com/wanggangkun/ST-Text-GCN}}\cite{cui2022self} utilizes a self-training method to extract keywords, ensuring the effective use of limited labeled text and many unlabeled texts.
\item HGAT\footnote{\url{https://github.com/BUPT-GAMMA/HGAT}}\cite{linmei2019heterogeneous} deploys a dual-level attention-based GNN to function on a corpus-level graph encompassing entities, topics, and documents.
\item SHINE\footnote{\url{https://github.com/tata1661/SHINE-EMNLP21}}\cite{wang2021hierarchical} introduces a hierarchically heterogeneous corpus-level graph that optimizes node interactions and captures textual similarities.
\end{enumerate} 

\noindent
\textbf{Group (C). Large Language Models}
 
Four LLMs were selected, with LLaMA2-7B and ChatGLM used for comparative experiments and the GPT-3 and FLAN-T5 series used in subsequent analytical experiments. 

\begin{enumerate}
\itemsep=0pt
\item LLaMA2-7B\footnote{\url{https://github.com/facebookresearch/llama-recipes/}}\cite{touvron2023llama}, introduced by Mata, encompasses seven billion parameters. It exhibits superior efficacy on multiple benchmark datasets and is suitable for research and commercial applications.
\item ChatGLM\footnote{\url{https://github.com/THUDM/ChatGLM-6B}}, introduced by Tsinghua University, is a robust language-generation model that provides advanced deep-learning technologies with training using extensive corpora.
\item GPT-3\cite{brown2020language}, note that GPT-3 does not release the model parameters, and we use them via the API. Consequently, the supervised fine-tuning of GPT-3 is not feasible; it is utilized solely for the experiments described in Section \ref{section4.7}.
\item FLAN-T5 series\footnote{\url{https://huggingface.co/google/flan-t5-xxl}}\cite{chung2022scaling}, introduced by Google. This method fine-tunes language models for tasks of an unprecedented scale owing to the remarkable generalization capacities of these models. Consequently, a singular model can effectively execute more than 1,000 tasks. In Section \ref{section4.8}, we perform comparative experiments using models of various sizes.
\end{enumerate} 

\subsection{Benchmark Comparison}
\begin{table*}
    \centering
    \resizebox{1.06\textwidth}{!}{
    \begin{tabular}{cccccccccccccc}
    \toprule
         &  & \multicolumn{2}{c}{MR}  & \multicolumn{2}{c}{Snipptes} & \multicolumn{2}{c}{Ohsumed} & \multicolumn{2}{c}{TagMyNews} & \multicolumn{2}{c}{StackOverflow} & \multicolumn{2}{c}{AGNews} \\
         \cmidrule(lr){3-4} \cmidrule(lr){5-6} \cmidrule(lr){7-8} \cmidrule(lr){9-10} \cmidrule(lr){11-12} \cmidrule(lr){13-14}
         &  & ACC & F1 & ACC & F1 &ACC & F1 &ACC & F1 &ACC & F1 &ACC & F1\\
    \midrule
         \multirow{3}{*}{PLMs} & BERT-avg & 51.69 & 50.65&79.31&78.47&23.91&4.98&55.13&44.26&$72.91^*$&$73.69^*$&$76.52^*$&$76.49^*$\\
         & BERT-CLS & 53.48&46.99&81.53&79.03&21.76&4.81&58.17&41.04&$73.74^*$&$74.11^*$&$78.35^*$&$78.42^*$ \\
         & RoBERTa &$53.62^*$&$52,27^*$&$79.58^*$&$79.10^*$&$26.95^*$&$19.47^*$&$55.57^*$&$50.45^*$&$64.87^*$&$64.40^*$&$79.33^*$&$79.45^*$ \\
    \midrule
         \multirow{5}{*}{GCNs} & HGAT-inductive& 61.18&59.77&79.40&77.69&42.08&25.71&58.20&49.55&$72.31^*$&$70.42^*$&70.23&68.43\\
         & SimpleSTC & 62.27&62.14&80.96&80.56&$43.16^*$&$23.35^*$&67.17&63.34&$73.63^*$&$73.39^*$&$72.62^*$&$71.89^*$\\
         & ST-Text-GCN & $50.23^*$&$34.02^*$&\underline{$83.83^*$}&\underline{$83.15^*$}&$33.64^*$&$22.62^*$&$52.33^*$&$47.38^*$&$69.68^*$&$68.94^*$&\underline{$86.83^*$}&\underline{$86.06^*$}\\
         & HGAT & 62.75&62.36&82.36&74.44&42.68&24.82&61.72&53.81&$75.29^*$&$75.14^*$&72.10&71.94\\
         & SHINE & 64.58&63.89&82.39&81.62&45.57&30.98&62.50&56.21&$76.81^*$&$76.44^*$&$81.39^*$&$81.45^*$\\
    \midrule
         \multirow{2}{*}{LLMs} &LLaMA2-7B& $71.49^*$&$71.03^*$&$78.47^*$&$78.76^*$&$48.08^*$&\underline{$40.21^*$}&$56.75^*$&$56.20^*$&\underline{$87.36^*$}&\underline{$88.21^*$}&$79.49^*$&$80.67^*$ \\
         & ChatGLM&\underline{$73.50^*$}&\underline{$73.31^*$}&$80.62^*$&$80.39^*$&\underline{$51.28^*$}&$36.68^*$&\underline{$70.77^*$}&\underline{$67.79^*$}&$86.13^*$&$86.77^*$&$82.51^*$&$82.50^*$\\
    \midrule
         \multirow{4}{*}{\textbf{Ours}}&\textbf{CDMT}& 76.10&76.10 & 84.67& 83.89 & 58.44 & 44.32 & 79.61 & 75.43 & 84.11 & 84.21 & 88.30 & 88.41\\
         &&\textcolor{red}{$\uparrow$} 2.60&\textcolor{red}{$\uparrow$} 2.88&\textcolor{red}{$\uparrow$} 0.84&\textcolor{red}{$\uparrow$} 0.74&\textcolor{red}{$\uparrow$} 7.16&\textcolor{red}{$\uparrow$} 4.11&\textcolor{red}{$\uparrow$} 8.84&\textcolor{red}{$\uparrow$}7.64&\textcolor{green}{$\downarrow$} 3.25&\textcolor{green}{$\downarrow$} 4.00&\textcolor{red}{$\uparrow$} 1.47&\textcolor{red}{$\uparrow$} 2.35
         \\&\textbf{SSE-CoT}&\textbf{81.70}&\textbf{81.72}&\textbf{85.75}&\textbf{85.32}&\textbf{61.10}&\textbf{51.85}&\textbf{83.37}&\textbf{79.84}&\textbf{89.67}&\textbf{89.58}&\textbf{89.14}&\textbf{89.28}
         \\ &&\textcolor{red}{$\uparrow$} 8.20&\textcolor{red}{$\uparrow$}8.41&\textcolor{red}{$\uparrow$}1.92&\textcolor{red}{$\uparrow$}2.17&\textcolor{red}{$\uparrow$} 9.82&\textcolor{red}{$\uparrow$} 11.64&\textcolor{red}{$\uparrow$} 12.60&\textcolor{red}{$\uparrow$}12.05&\textcolor{red}{$\uparrow$} 2.31&\textcolor{red}{$\uparrow$} 1.37&\textcolor{red}{$\uparrow$} 2.31&\textcolor{red}{$\uparrow$} 3.22
         \\
    \bottomrule
    \end{tabular}
    }
    \caption{Performance evaluation measured on short text datasets. This table presents the comparative performance of baseline models and ours, measured in terms of ACC (\%) and F1 (\%). The table highlights the highest scores in bold, with underscores indicating the highest scores achieved by prior methods. $*$ indicates that the result is reproduced by us.}
    \label{table2}
\end{table*}

Table \ref{table2} shows the performance comparison. As can be seen, our SSE-CoT method surpasses other approaches on all six datasets. Similarly, the CDMT method shows notable performance across five datasets. Specifically, on the TagMyNews dataset, SSE-CoT achieves an increase of 12.60\% in ACC and 12.05\% in F1 score compared to the previously optimal ChatGLM model. CDMT records an 8.84\% improvement in ACC and a 7.64\% increase in F1 score. These results support our hypothesis that LLMs can effectively utilize their inherent knowledge and abilities to address traditional NLP tasks, particularly the STC tasks discussed in this paper, thereby validating the effectiveness of our proposed methods.

A comparative analysis revealed that SSE-CoT outperformed CDMT, especially on the MR dataset, where SSE-CoT's accuracy of SSE-CoT exceeded that of CDMT by 5.6\%. This indicates that models with larger parameters have greater intrinsic knowledge and enhanced capabilities to solve the STC tasks. Although the CML did not outperform ChatGLM on the StackOverflow dataset, it significantly exceeded the top-performing model in the GCN-based group. Moreover, it achieved a 7.3\% higher ACC and a 7.77\% greater F1 score than SHINE.

An interesting observation is that for the Snippets and AGNews datasets, the methods based on LLMs did not outperform those utilizing GCNs. In the Snippets dataset, ST-Text-GCN outperformed ChatGLM by 3.21\% in ACC and 2.76\% in F1 score. Given the dense entity relationship characteristics of news datasets, the inherent topological advantages of GCN methods can enable a more effective capture of relational data, leading to superior performance. Consequently, traditional approaches have retained their relevance in the context of LLMs.
\subsection{Ablation Study}

\begin{table*}
    \centering
    \begin{tabular}{ccccccccccccc}
    \toprule
     & \multicolumn{2}{c}{MR}  & \multicolumn{2}{c}{Snipptes} & \multicolumn{2}{c}{Ohsumed} & \multicolumn{2}{c}{TagMyNews} & \multicolumn{2}{c}{StackOverflow} & \multicolumn{2}{c}{AGNews} \\
         \cmidrule(lr){2-3} \cmidrule(lr){4-5} \cmidrule(lr){6-7} \cmidrule(lr){8-9} \cmidrule(lr){10-11} \cmidrule(lr){12-13}
        & ACC & F1 & ACC & F1 &ACC & F1 &ACC & F1 &ACC & F1 &ACC & F1\\
    \midrule
        CDMT & \textbf{76.10}& \textbf{76.10} & \textbf{84.67}& \textbf{83.89} & \textbf{58.44}	& \textbf{44.32} & \textbf{79.61} & \textbf{75.43} & \textbf{84.11} & \textbf{84.21} & \textbf{88.30} & \textbf{88.41} \\
      w/o ECCA & 75.48&75.46&83.27&81.80	&56.91	&48.65&77.07&	74.06&	83.08&	83.44&	87.52&	87.74\\
       w/o SSE-CoT  & 74.36&74.38	&81.55&80.64&56.43&	47.96&75.96&72.19&82.87&83.58&86.38&86.80 \\
       w/o DA-CoT& 74.02&74.02&81.21&80.17&55.18&47.27	&75.45	&71.73	&81.96&	82.74&85.87	&86.10 \\
    \bottomrule
    \end{tabular}
    \caption{Performance evaluation of CDMT and its variant on short text datasets. The best results are in bold. `w/o ECCA' illustrates results without employing the ECCA strategy, while `w/o SSE-CoT' and `w/o DA-CoT' display outcomes when omitting SSE-CoT and DA-CoT rationales, respectively.}
    \label{table3}
\end{table*}

\noindent
\textbf{CDMT Ablation.} Ablation studies were performed on six benchmark datasets to assess the influence of particular strategies or components on the CDMT method. We developed three variations of the CDMT, as follows:
\begin{itemize}
    \itemsep=0pt
    \item 
    CDMT+w/o ECCA: During the fine-tuning stage, the original text $X$ is used as input to predict the label without employing the ECCA strategy.
    \item 
    CDMT+w/o SSE-CoT: The task of generate SSE-CoT rationales is removed from the multi-task learning process.
    \item 
    CDMT+w/o DA-CoT: Similarly, the task of generate DA-CoT rationales is excluded from the multi-task learning process.
\end{itemize}

Table \ref{table3} displays the outcomes of our ablation study, highlighting the best-performing metrics in bold. These findings demonstrate the integral role of each strategy and component in the effectiveness of our CDMT method. Notably, ECCA contributes most substantially, with SSE-CoT and DA-CoT also providing significant enhancements. The significant improvement provided by ECCA confirms the importance of a prompt for generative models with an encoder-decoder architecture. The improvement observed with SSE-CoT and DA-CoT affirms that our strategy effectively transfers the knowledge and capabilities of LLM to a smaller model.

The impact of SSE-CoT and DA-CoT varied according to the dataset type. For instance, in news-related datasets, such as TagMyNews, SSE-CoT exerts a more substantial influence than DA-CoT, likely because of the frequent occurrence of widely recognized entities in news texts. Conversely, in specialized datasets, such as Ohsumed for medical content and StackOverflow for computing science, the importance of DA-CoT increases, reflecting the prevalence of domain-specific terminology.

\begin{table*}
    \centering
    \begin{tabular}{ccccccccccccc}
    \toprule
     & \multicolumn{2}{c}{MR}  & \multicolumn{2}{c}{Snipptes} & \multicolumn{2}{c}{Ohsumed} & \multicolumn{2}{c}{TagMyNews} & \multicolumn{2}{c}{StackOverflow} & \multicolumn{2}{c}{AGNews} \\
         \cmidrule(lr){2-3} \cmidrule(lr){4-5} \cmidrule(lr){6-7} \cmidrule(lr){8-9} \cmidrule(lr){10-11} \cmidrule(lr){12-13}
        & ACC & F1 & ACC & F1 &ACC & F1 &ACC & F1 &ACC & F1 &ACC & F1\\
    \midrule
        SSE-CoT & \textbf{81.70}& \textbf{81.72} & \textbf{85.75}& \textbf{85.32} & \textbf{61.10}& \textbf{51.85} &  \textbf{83.37} & \textbf{79.84} & \textbf{89.67} & \textbf{89.58}& \textbf{89.14} & \textbf{89.28} \\
      w/o rewriting & 81.38&81.38	&85.28&84.16	&60.80&	49.63&82.78&78.91&89.08&88.44&	88.52&	89.04\\
       w/o retrieval & 80.18&80.20	&83.65&83.72&	57.43&42.96&	81.04&76.48&87.87&87.58&87.30&88.61 \\
       w/o both & 79.84&79.84&83.16&83.51&56.99&40.65 &80.44&76.12&87.35&87.51&87.02&	87.94\\
    \bottomrule
    \end{tabular}
    \caption{Performance evaluation of SSE-CoT and its variant on short text datasets. The best results are in bold. `w/o rewriting' and `w/o retrieval' refer to results without the rewriting step and retrieval step in SSE-CoT, respectively. ` w/o both' refers to the exclusion of both steps}
    \label{table4}
\end{table*}

\noindent
\textbf{SSE-CoT Ablation.} Ablation studies were performed on six benchmark datasets to assess the influence of the components of the SSE-CoT framework. We developed two variations of the QLFQ as follows.
\begin{itemize}
    \itemsep=0pt
    \item 
    SSE-CoT+w/o rewriting: Omission of the rewriting step, using the concatenation of the original input with the retrieved text as input.
    \item 
    SSE-CoT+w/o retrieval: Omission of the retrieval step, employing rewritten original inputs as input.
    \item 
    SSE-CoT+w/o both: Utilizing the original input without modification.
\end{itemize}

Table \ref{table4} displays the outcomes of our ablation study, highlighting the best-performing metrics in bold. The findings indicate that both steps in our SSE-CoT are beneficial. The rewriting step addresses the challenge of syntactic inexactitude in short texts, while the retrieval step effectively resolves the issue of semantic sparsity. A comparison reveals that retrieval offers slightly more advantage than rewriting, suggesting that semantic deficiencies in short texts are more critical than syntactic imprecision.

\subsection{Analysis of In-Context Learning for SSE-CoT}\label{section4.7}
\begin{figure*}
    \centering
    \includegraphics[width=1\textwidth]{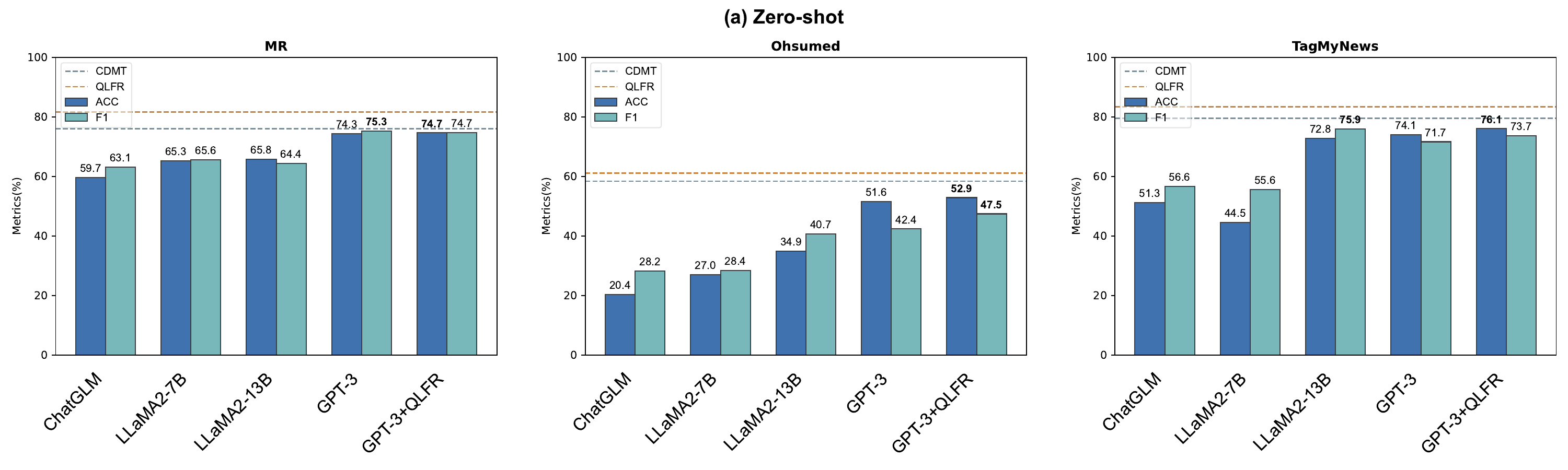}
    \includegraphics[width=1\textwidth]{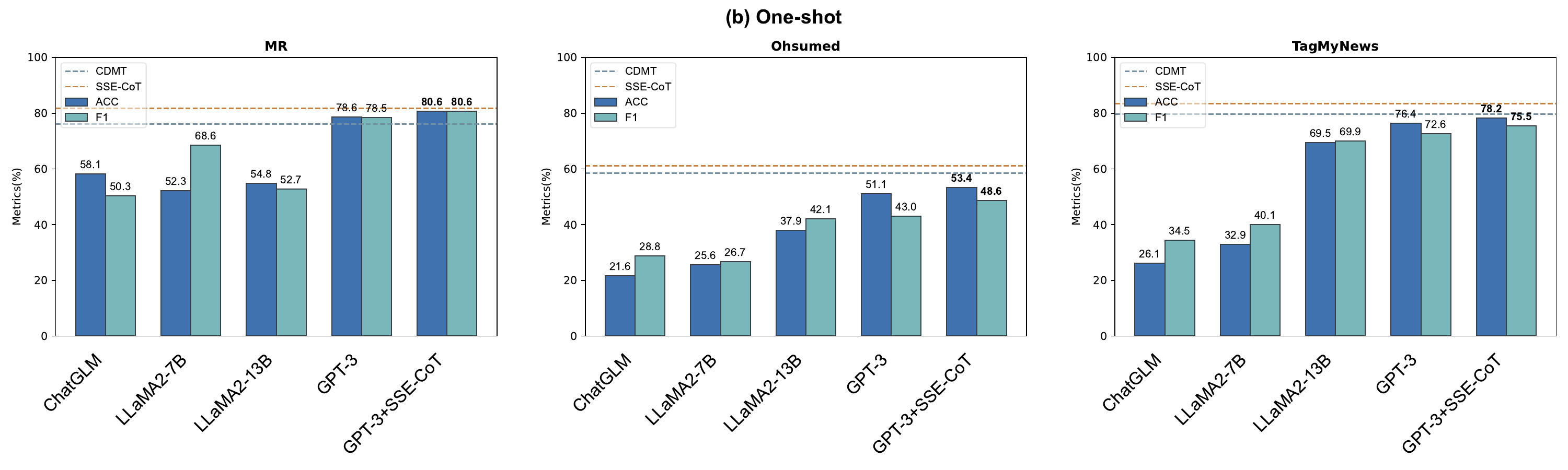}
    \caption{Performance evaluation of LLMs in zero-shot and one-shot settings is conducted using three representative datasets. The upper three groups correspond to zero-shot settings, while the lower three pertain to one-shot settings. In each figure, the best results are highlighted in bold.}
    \label{fig5}
\end{figure*}

In addition to the supervised fine-tuning (SFT) paradigm, in-context learning has gained popularity. Four LLMs, ChatGLM, LLaMA2-7B, LLaMA2-13B, and GPT-3 were selected for experimentation under the zero-shot and one-shot settings. Given the rigorous demands of in-context learning, the SSE-CoT was tested solely on the GPT-3. We manually selected a sample from each category in a one-shot setting and constructed the SSE-CoT as the context. Specific examples of the input formats are provided in Appendix \ref{appendixB}. The experimental results are illustrated in Fig \ref{fig5}.

\noindent
\textbf{Zero-shot setting.} 
In the zero-shot setting analysis conducted across three datasets, the in-context learning abilities of the four models were ranked as follows: GPT-3 outperformed LLaMA2-13B, which, in turn, surpassed both ChatGLM and LLaMA2-7B. This hierarchy can be attributable to in-context learning capabilities inherent in larger models, with increased parameters correlating with enhanced performance. Applying our SSE-CoT method to GPT-3 resulted in performance gains across the board. For instance, the ACC on the Ohsumed dataset increased from 51.6\% to 52.9\%. These findings indicate that our approach is not limited to the SFT but is also applicable to the context learning paradigm.

\noindent
\textbf{One-shot setting.} In the one-shot setting, the findings regarding the model performance were consistent with those observed in the zero-shot setting. Furthermore, our method achieved improvements across all datasets.

Comparing the results of the zero-shot and one-shot tasks, contrary to our initial expectations, the hypothesis that providing task-relevant examples enhances the comprehension and task response of LLMs was not supported. The findings indicate that only GPT improves performance in one-shot learning compared with zero-shot learning on two of the datasets. In contrast, a decline is observed in the remaining datasets. Further analysis of the model outputs suggested that LLM with approximately 10 billion parameters tended to demonstrate a reduced capacity for processing instructions with greater input complexity. In contrast, the GPT consistently exhibits a robust understanding of instructions and sustains its performance despite increased context length.

Comparison of the results of SFT and in-context learning paradigm. GPT-3, with over a hundred billion parameters, exhibits a remarkable in-context learning performance. In the one-shot setting, GPT-3 exceeded CDMT, and GPT-3+SSE-CoT nearly matched the supervised SSE-CoT method. However, for models with billions of parameters, the SFT paradigm remained predominant.

\subsection{Analysis of different base models for CDMT}
In our CDMT method, the initial stage involves the extraction of rationales with the assistance of an LLM, followed by a fine-tuning phase that requires a smaller model. Therefore, we investigated the outcomes of employing the same LLM with various smaller models and the results of utilizing the same smaller model with different LLMs.
\begin{figure*}
    \centering
    \includegraphics[width=0.85\textwidth]{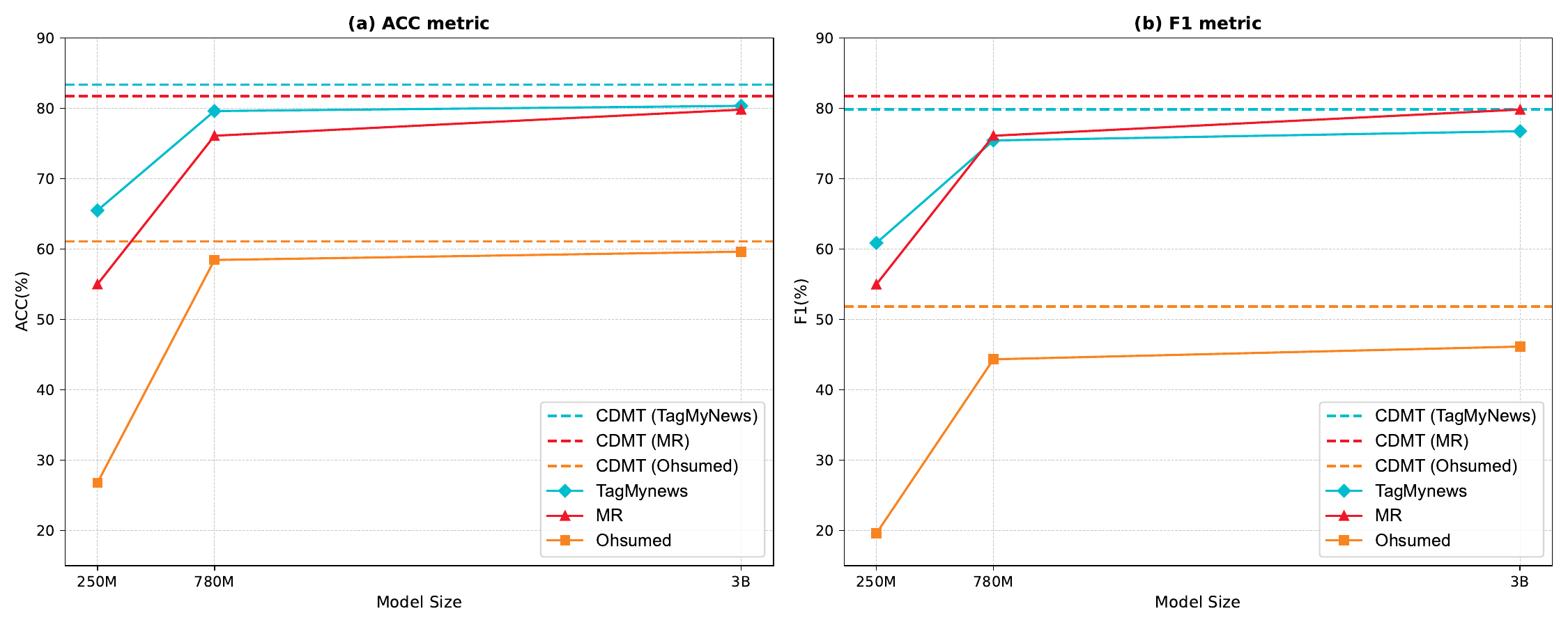}
    \caption{Evaluation of performance across various smaller model sizes reveals that as the parameter count in these models escalates, there is a performance improvement, albeit not in direct proportion.}
    \label{fig6}
\end{figure*}

\noindent
\textbf{Different smaller models.} As shown in Fig \ref{fig6}, we utilized LLaMA2-13B as the LLM and three versions of Flan-T5-Base, Flan-T5-Large, and Flan-T5-XL as smaller models, with parameter counts of 250M, 780M, and 3 B, respectively. It is apparent that, with an increase in the parameter size of the smaller models, there is a corresponding enhancement in both the ACC and F1 metrics, which suggests a correlation between the capacity of the smaller model and its performance. However, this relationship is not strictly linear. For instance, in the Ohsumed dataset, the increase in ACC from the 250M model to the 780M model does not reflect the expected proportional improvement, indicating diminishing returns as the model size increases.

\begin{table}[H]
    \centering
    \resizebox{\linewidth}{!}{
    \begin{tabular}{ccccccc}
    \toprule
    & \multicolumn{2}{c}{MR}  &  \multicolumn{2}{c}{Ohsumed} & \multicolumn{2}{c}{TagMyNews} \\
    \cmidrule(lr){2-3} \cmidrule(lr){4-5} \cmidrule(lr){6-7} 
    & ACC & F1 & ACC & F1 &ACC & F1 \\
    \midrule
        ChatGLM& 70.17&	70.54	&55.11	&47.25	&73.75&	70.48 \\
        LLaMA2-7B & 73.89	&73.89	&55.90	&49.03	&77.63	&74.09 \\
        Flan-T5-XXL & 72.44	&72.65&	51.72	&44.10	&76.71&	72.97\\
        LLaMA2-13B & 76.10	&76.10&	58.44&	44.32&	79.61	&75.43\\
        GPT-3 & \textbf{79.46}&\textbf{79.46}	&\textbf{59.51}&\textbf{46.85}&	\textbf{81.07}&\textbf{77.95}\\
    \bottomrule
    \end{tabular}}
    \caption{Evaluation of performance across various-sized Large Language Models: Larger models demonstrate enhanced knowledge transfer capabilities attributable to their expanded parameter count.}
    \label{table5}
\end{table}

\noindent
\textbf{Different LLMs.} We selected five models as representatives of LLMs and Flan-T5-Large as a comparatively smaller model. The experimental results presented in Table \ref{table5} indicate that GPT-3, as an LLM, transfers knowledge and capabilities most effectively to Flan-T5-Large. On the MR dataset, GPT-3's ACC and F1 scores exceed those of LLaMA2-13B by 3.36\% and 3.36\%, respectively. Despite having fewer parameters than Flan-T5-XXL, LLaMA2-7B outperformed it on the Ohsumed and TagMyNews datasets, which may be attributed to differences in the model's training corpus and strategies.

\subsection{Analysis of different prompts for CDMT}\label{section4.8}
To verify the efficacy of the proposed ECCA, we conducted comparative experiments using two prompts with different semantic richness. Details of the prompts are provided in the Appendix\ref{appendixC}.

\begin{table}[H]
    \centering
    \resizebox{\linewidth}{!}{
    \begin{tabular}{ccccccc}
    \toprule
    & \multicolumn{2}{c}{MR}  &  \multicolumn{2}{c}{Ohsumed} & \multicolumn{2}{c}{TagMyNews} \\
    \cmidrule(lr){2-3} \cmidrule(lr){4-5} \cmidrule(lr){6-7} 
    & ACC & F1 & ACC & F1 &ACC & F1 \\
    \midrule
        w/o prompt &75.48&75.46&56.91&48.65&77.07&74.06\\
        ours& 76.10&76.10&\textbf{58.44}&44.32&\textbf{79.01}&\textbf{75.43}\\
        prompt1 & 75.29	&75.30&57.28&48.33&77.62&74.68 \\
        prompt2 &\textbf{76.15} &\textbf{76.15}&58.26&\textbf{48.71}&78.89&75.36\\
    \bottomrule
    \end{tabular}}
    \caption{Evaluation of Different Prompts in CDMT: the efficacy of ECCA demonstrated through significant outcomes.}
    \label{table6}
\end{table}

Table \ref{table6} highlights several interesting trends. Prompts enhanced performance across all measured metrics relative to their absence. Furthermore, our method outperformed the less semantically rich prompt1 and demonstrated greater accuracy than prompt2 in the two datasets. Although our approach does not outperform prompt2 on the MR dataset in terms of ACC, it offers considerable benefits. It obviates the necessity for creating manual, dataset-specific prompts, which yield notably shorter input lengths than prompt2 by simply appending label names, benefiting processing efficiency and model scalability.

\subsection{Analysis of training size}
We investigated the impact of the training data ratio on the model performance. Experiments were conducted using SHINE, ChatGLM, and our proposed methods on the MR and TagMyNews datasets.
\begin{figure*}
    \centering
    \includegraphics[width=0.85\textwidth]{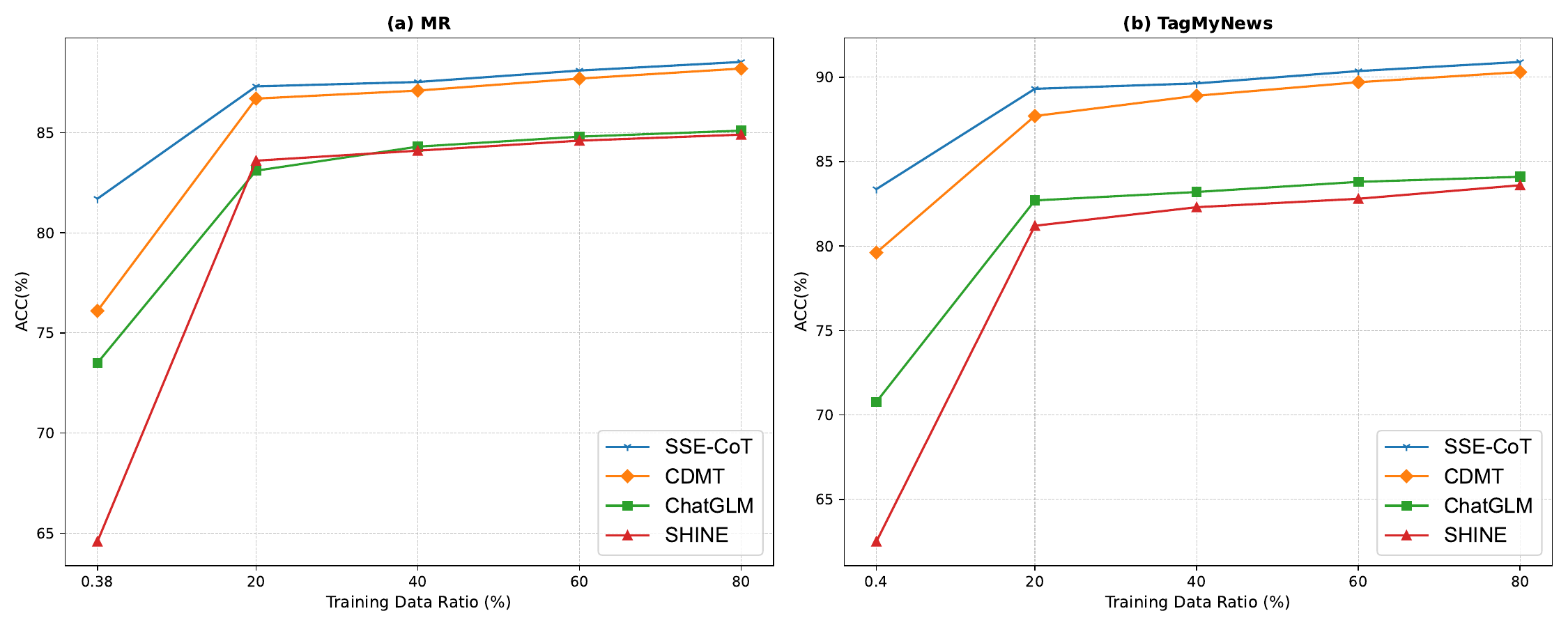}
    \caption{Performance evaluation of different ratios of training data.}
    \label{fig7}
\end{figure*}

The results presented in Fig \ref{fig7} show that the proposed method significantly outperforms the other two models in low-resource environments. The data reveal a consistent increase in test accuracy correlated with the rising proportion of training data for both the MR and TagMyNews datasets, consistent with supervised learning principles. Additionally, the acceleration or plateau in ACC improvement suggests diminishing returns in the model performance beyond a certain data threshold. Notably, the ACC of SHINE and ChatGLM approach parity, whereas our methods consistently outperform them, indicating that, with adequate data, the inherent strengths of the model become increasingly influential.

\subsection{Analysis of time complexity}
The results presented in Table \ref{table7} assess the time complexity of four methods using the TagMyNews dataset, which includes a training set of 140 samples and a test set of 24,600 samples. KG denotes constructs a knowledge graph and RG stands for rationale generation. As shown, SHINE constructs a knowledge graph in 340 seconds, whereas CDMT generates rationales using a large language model in only 13 seconds. In the training phase, SHINE completes its process in 94 seconds, while SSE-CoT requires significantly more time at 1859 seconds. For inference time, CDMT and SSE-CoT require 326 seconds and 804 seconds, respectively.

Traditional methods such as SHINE excel in training speed compared to LLM-based approaches. However, in practical applications, inference time is crucial. SHINE requires reconstruct and retrain each new test sample, making its inference time equal to its construction and training times combined. In contrast, CDMT's pipeline design is more efficient, offering considerable advantages. While SSE-CoT inference is slower than SHINE, it remains feasible for large-scale deployment.
\begin{table}[H]
    \centering
    \resizebox{0.85\linewidth}{!}{
    \begin{tabular}{cccc}
    \toprule
        &RG/KG(s) &Train(s)  &  Inference(s) \\
    \midrule
        SHINE &340&94&434\\
        ChatGLM&\diagbox{}{} &1446&752\\
        CDMT &13& 577	&326\\
        SSE-CoT &\diagbox{}{}  &1859	&804\\
    \bottomrule
    \end{tabular}}
    \caption{Evaluation of time complexity on TagMyNews Dataset.}
    \label{table7}
\end{table}

\section{Conclusion}
In this study, we developed and evaluated novel methods to improve Short Text Classification (STC) using Large Language Models (LLMs) and a Chain-of-Thought (CoT) processing approach. The Semantic and Syntactic Enrichment CoT (SSE-CoT) method breaks down the STC tasks into four steps, facilitating thorough comprehension and management of short texts. This approach outperforms traditional models by providing a level of semantic and syntactic analysis that was not achievable with Graph Convolutional Networks (GCNs). In parallel, acknowledging the challenges faced in resource-constrained sectors like finance and healthcare, we introduced the CoT-Driven Multi-Task learning (CDMT) framework. This method leverages insights from LLMs and adapts them for smaller models, improving their efficiency and effectiveness through targeted fine-tuning and multi-task learning strategies. Comprehensive experiments were conducted across six prevalent datasets to evaluate the effectiveness of the proposed methods. Experimental results indicate that the proposed methods significantly outperformed established baselines. However, this complex task remains partially unresolved because its performance on domain-specific datasets remains suboptimal. Future research should prioritize the integration of LLMs with additional knowledge sources to further refine the proposed methodologies.

\section{Limitations}
Our study validated the effectiveness of the proposed method through carefully designed experiments, although several limitations were encountered. Constraints related to hardware resources and time restricted the number of LLMs employed in this study. Moreover, the rapid advancements in LLMs  technologies pose challenges in maintaining up-to-date comparative analyses. A significant limitation is the increased time complexity during the training and inference phases of our methods, a typical trade-off when using advanced models. Despite these challenges, we made considerable efforts to ensure the robustness and relevance of our findings. In future work, we aim to expand our research by incorporating more diverse LLMs and enhancing the efficiency of our algorithms. This approach seeks to address the current limitations related to hardware and time constraints while enhancing the scalability of our methods. Additionally, we intend to update our comparative analyses regularly to align with the rapid advancements in LLMs technologies.

\bibliographystyle{unsrt}
\bibliography{cas-refs}

\newpage
\appendix

\section{DA-CoT: Cross-Domain Applications}\label{appendixA}
\begin{figure*}[h]
    \centering
    \includegraphics[width=0.85\textwidth]{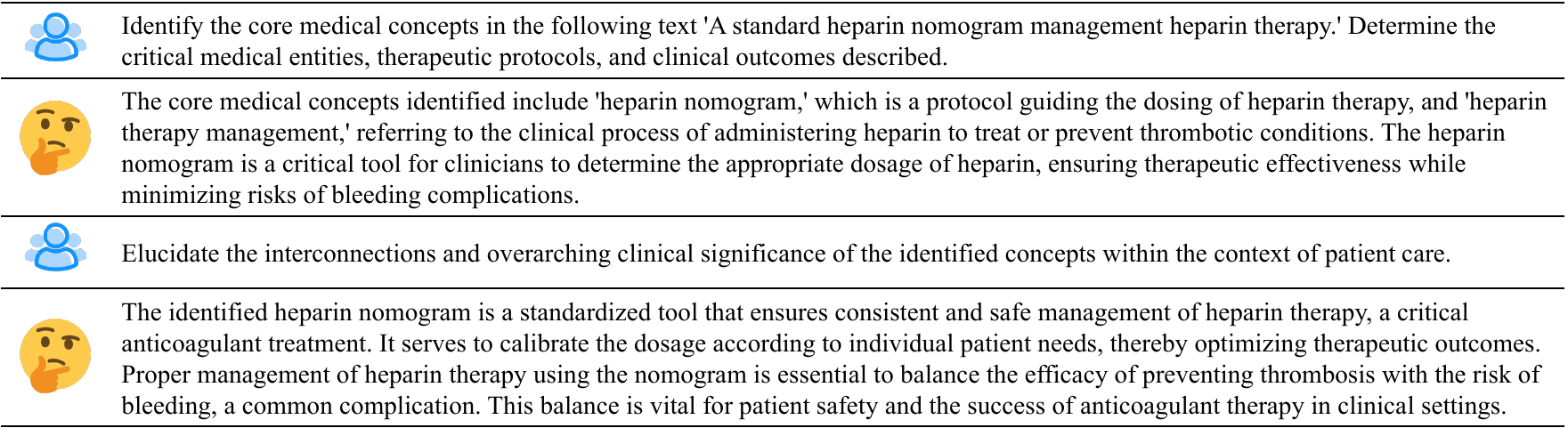}
    \caption{DA-CoT of medical domain}
\end{figure*}
\begin{figure*}[h]
    \centering
    \includegraphics[width=0.85\textwidth]{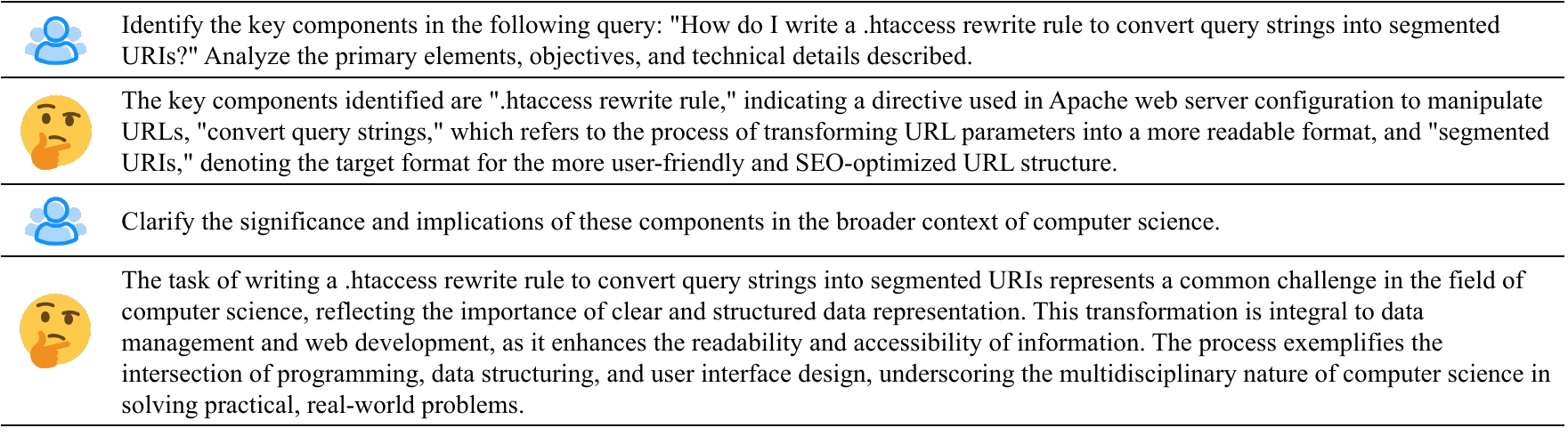}
    \caption{DA-CoT of computer science domain}
\end{figure*}

\newpage
\section{In-Context Learning input apllications}\label{appendixB}
\begin{figure*}
    \centering
    \includegraphics[width=0.65\textwidth]{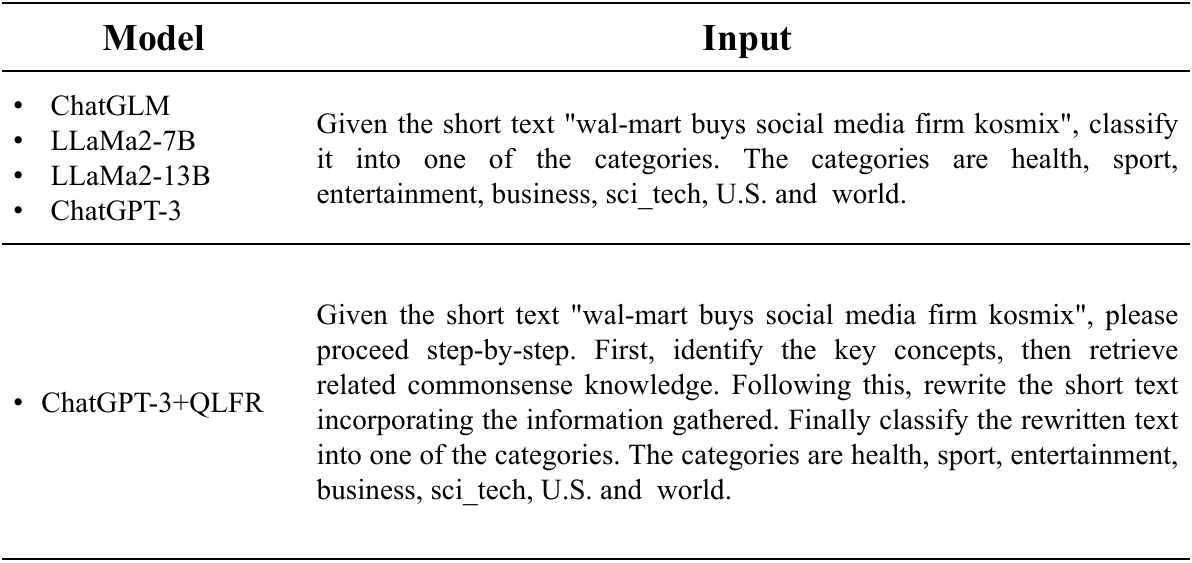}
    \caption{zero-shot setting}
\end{figure*}
\begin{figure*}
    \centering
    \includegraphics[width=0.9\textwidth]{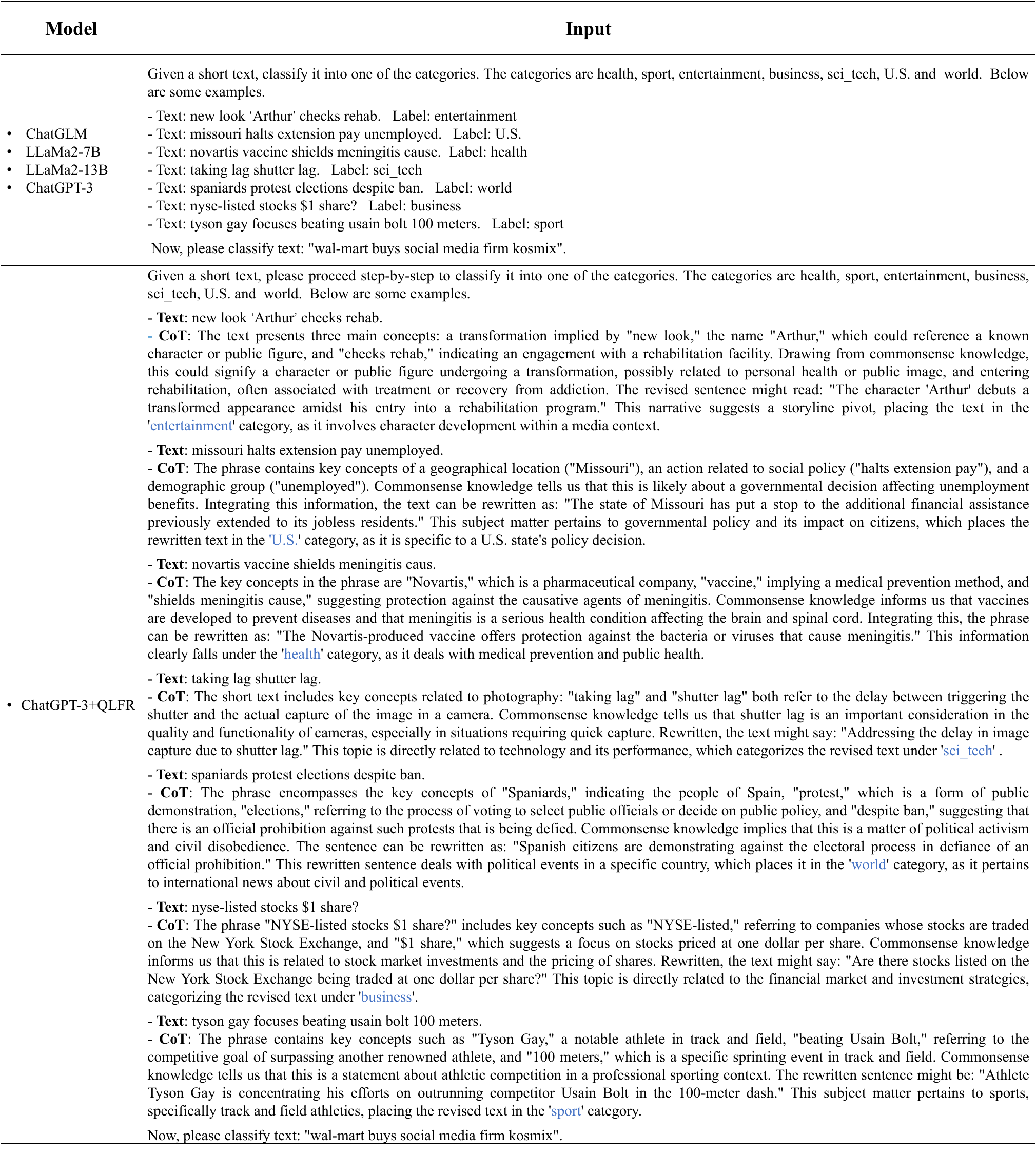}
    \caption{one-shot setting}
\end{figure*}

\section{Different Prompts for CDMT}\label{appendixC}
\begin{mdframed}[
    frametitle={Prompt1},
    frametitlerule=true,
    frametitlebackgroundcolor=gray!60,
    linewidth=1.5pt,
    roundcorner=5pt, 
    backgroundcolor=gray!10, 
    linecolor=black, 
    align=center, 
    userdefinedwidth=0.45\textwidth,
]
\textsf{Categorize this text: `wal-mart buys social media firm kosmix'.}
\end{mdframed}

\begin{mdframed}[
    frametitle={Prompt2},
    frametitlerule=true,
    frametitlebackgroundcolor=gray!60,
    linewidth=1.5pt,
    roundcorner=5pt, 
    backgroundcolor=gray!10, 
    linecolor=black, 
    align=center, 
    userdefinedwidth=0.45\textwidth,
]
\textsf{Given the short text `wal-mart buys social media firm kosmix', classify it into one of the categories. The categories are health, sport, entertainment, business, sci\_tech, U.S. and  world.}
\end{mdframed}

\section{Case study}
\begin{figure*}
    \centering
    \includegraphics[width=1\textwidth]{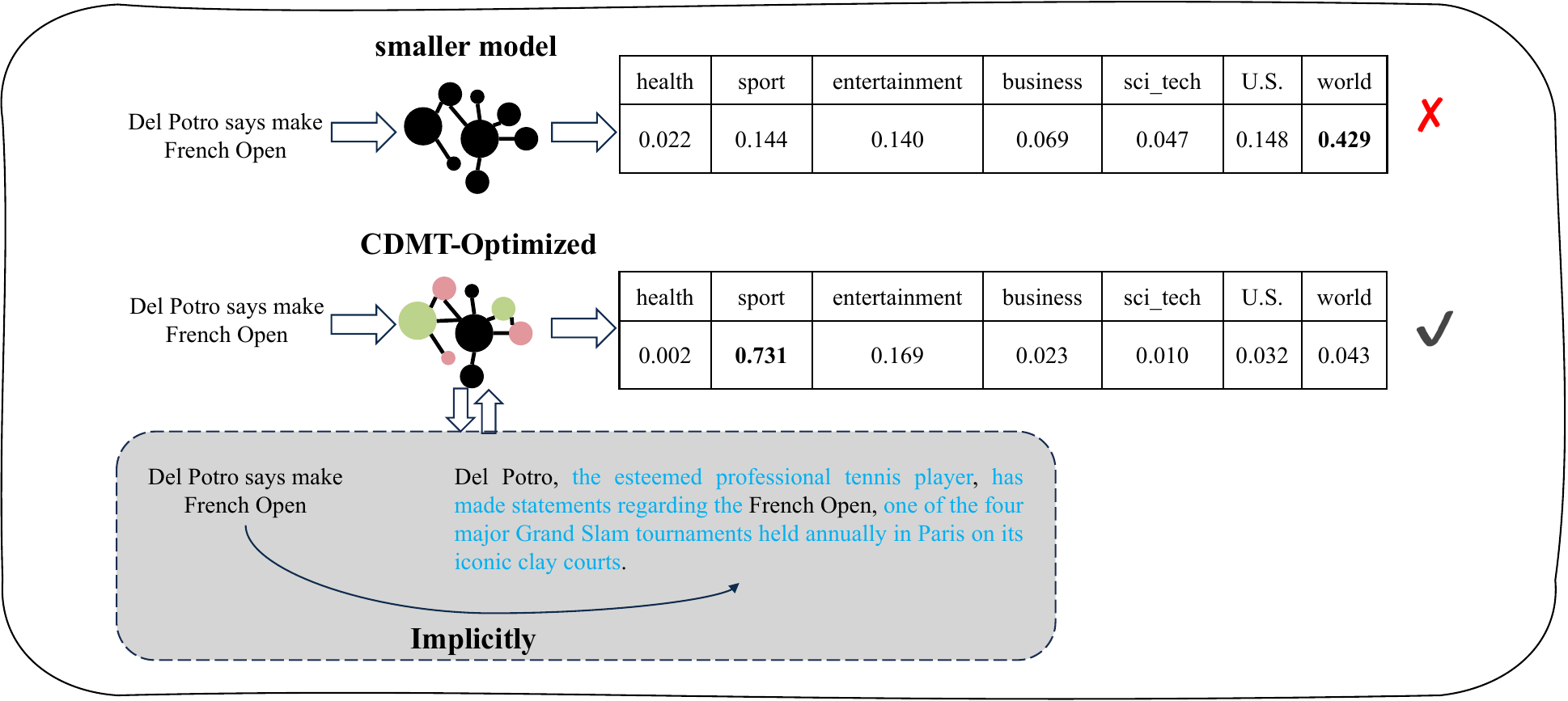}
    \caption{case study of CDMT}
    \label{fig12}
\end{figure*}

Figure \ref{fig12} presents a case study where our CDMT framework optimized a smaller model to enhance short text classification. Initially, the model incorrectly classified the phrase `Del Potro says make French Open' as `world' due to poor semantic and syntactic comprehension. Following optimization, the model's accuracy improved markedly, correctly categorizing the phrase under `sports'. This enhancement illustrates the impact of advancing the model's understanding of semantics and syntax on classification accuracy and reliability.

\end{document}